\documentclass[preprint,12pt]{elsarticle}

\usepackage{amssymb}
\usepackage{lipsum}
\usepackage{graphicx} 
\usepackage{subcaption}
\usepackage{hyperref}
\usepackage{booktabs}
\usepackage{xr}
\usepackage{xr-hyper}
\usepackage{hyperref}
\usepackage{xcolor}

\newlength{\subfigwidth}
\newlength{\subfigexamplewidth}
\newcommand{\ready}{\textcolor[rgb]{0.0,0.0,0.0}}

\usepackage{lineno}

\journal{Smart Agricultural Technology}

\begin{document}

\begin{frontmatter}



\title{\ready{Automated {R}e-{I}dentification of {H}olstein-{F}riesian Cattle in Dense Crowds}}


\author[inst1]{Phoenix Yu\corref{cor1}}
\ead{ho19002@bristol.ac.uk}
\author[inst1]{Tilo Burghardt}
\author[inst2]{Andrew W Dowsey}
\author[inst1]{Neill W Campbell}

\affiliation[inst1]{organization={School of Computer Science},
            addressline={Merchant Venturers Building, Woodland Road, University of Bristol}, 
            city={Bristol},
            postcode={BS8 1UB}, 
            state={Bristol},
            country={United Kingdom}}


\affiliation[inst2]{organization={Bristol Veterinary School, University of Bristol},
            addressline={Langford House, Dolberry, Churchill}, 
            city={Bristol},
            postcode={BS40 5DU}, 
            state={Bristol},
            country={United Kingdom}}

\cortext[cor1]{Corresponding Author, PhD Student}


\begin{abstract}
\ready{Holstein-Friesian detection and re-identification (Re-ID) methods capture individuals well when targets are spatially separate. However, existing approaches, including YOLO-based species detection, break down when cows group closely together. This is particularly prevalent for species which have outline-breaking coat patterns. To boost both effectiveness and transferability in this setting, we propose a new detect-segment-identify pipeline that leverages the Open-Vocabulary Weight-free Localisation and the Segment Anything models as pre-processing stages alongside Re-ID networks. To evaluate our approach, we publish a collection of nine days CCTV data filmed on a working dairy farm. Our methodology overcomes detection breakdown in dense animal groupings, resulting in a $98.93\%$ accuracy. This significantly outperforms current oriented bounding box-driven, as well as SAM species detection baselines with accuracy improvements of $47.52\%$ and $27.13\%$, respectively. We show that unsupervised contrastive learning can build on this to yield $94.82\%$ Re-ID accuracy on our test data. Our work demonstrates that Re-ID in crowded scenarios is both practical as well as reliable in working farm settings with no manual intervention. Code and dataset are provided\footnote{\href{https://phoenix4582.github.io/dazzlecows.github.io/}{https://phoenix4582.github.io/dazzlecows.github.io/}} for reproducibility.}

\end{abstract}



\end{frontmatter}


\let\clearpage\relax
\section{Introduction}
\label{sec:introduction}
\begin{figure}[!htbp]
\centering
\captionsetup{justification=centering}
\begin{tabular}{p{0.49\textwidth} p{0.49\textwidth}}
\begin{subfigure}[b]{0.49\textwidth}
    \includegraphics[width=\linewidth]{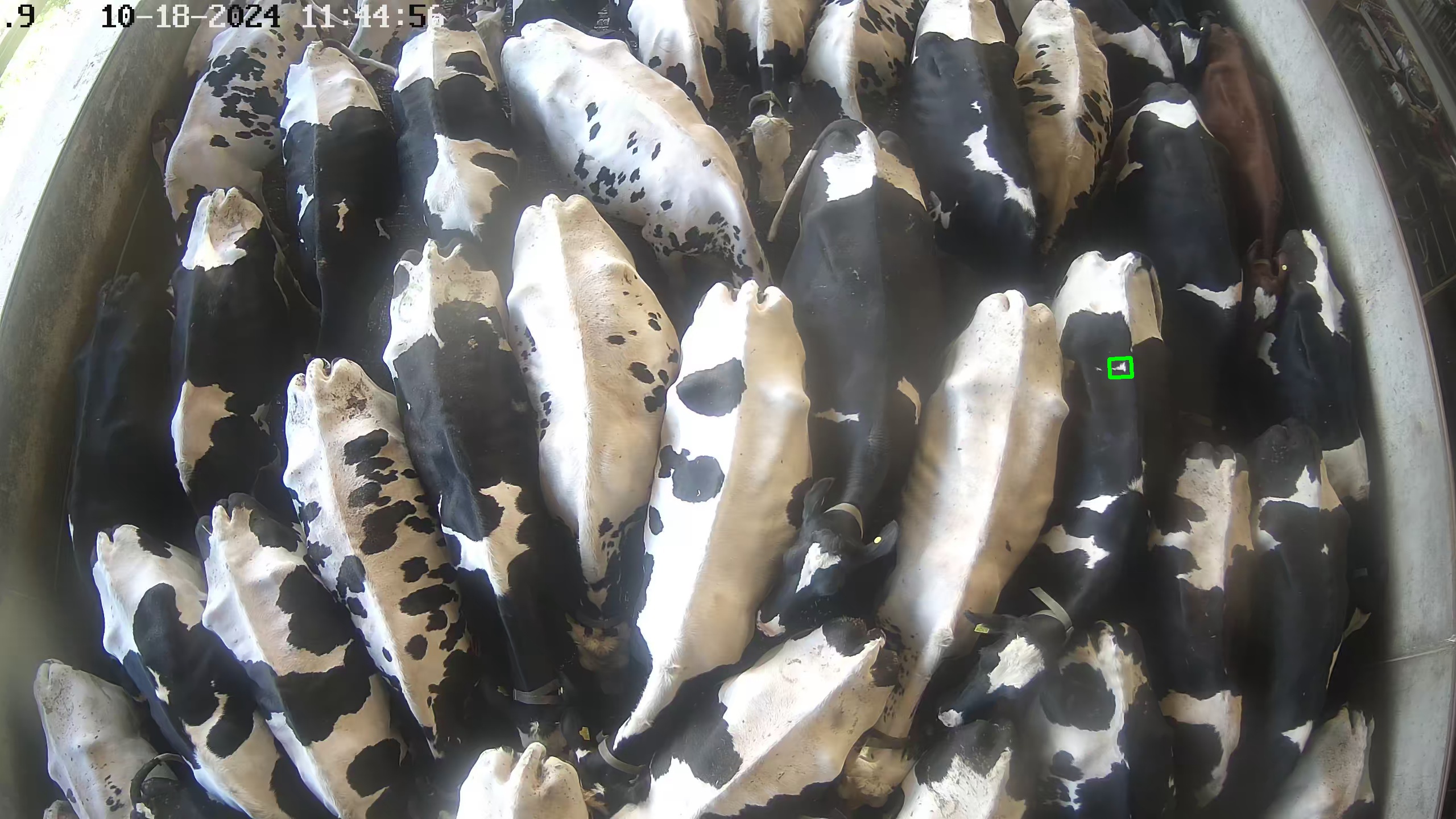}
    \caption{YOLO-v11x}
\end{subfigure} &
\begin{subfigure}[b]{0.49\textwidth}
    \includegraphics[width=\linewidth]{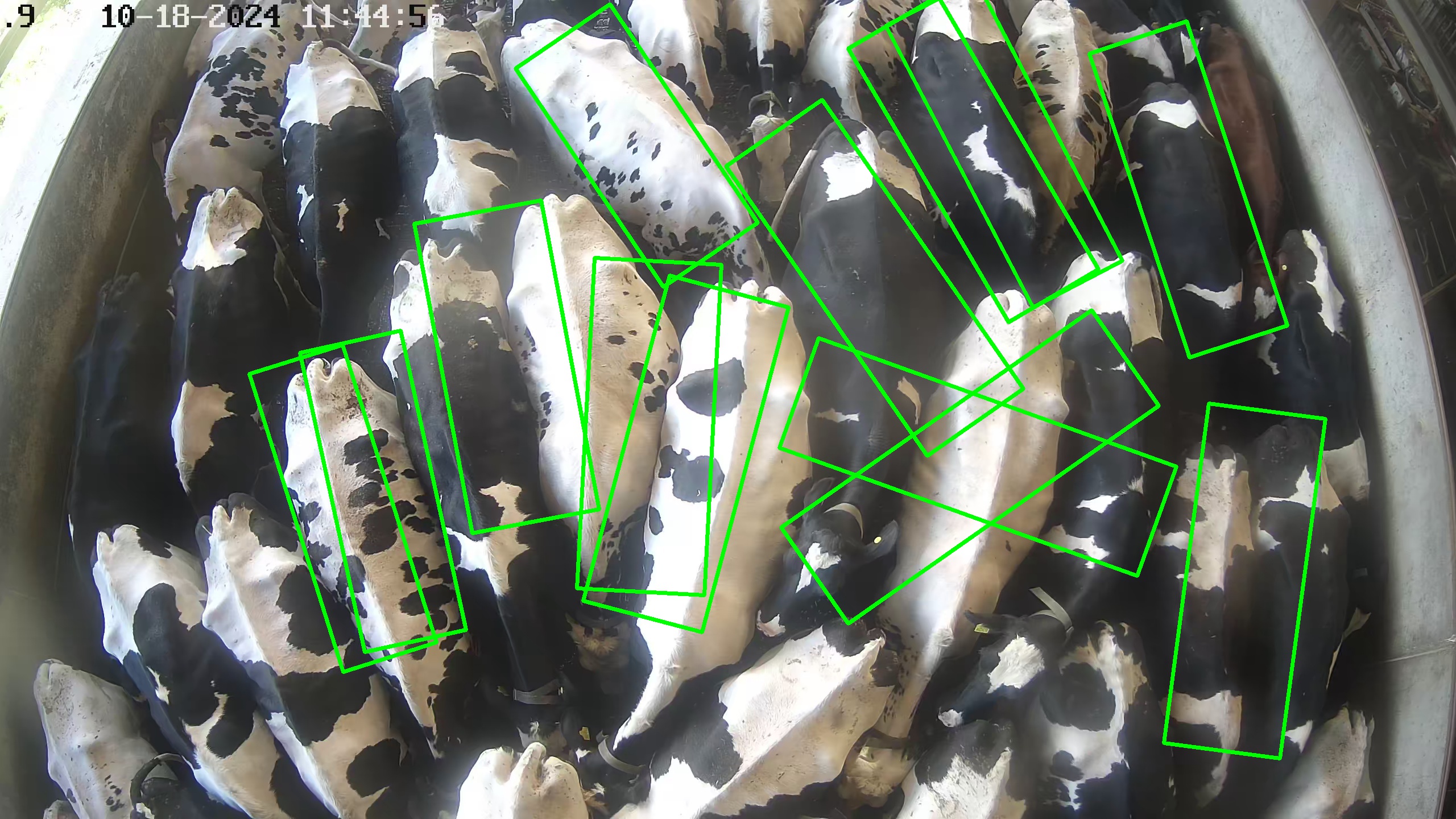}
    \caption{RetinaNet}
\end{subfigure}
\end{tabular}
\caption{\ready{\textbf{Existing Detector Performance on Crowded Cows.} Visualisations of performance with YOLO-v11x~\cite{khanam2024yolov11}~\textit{(left)} and RetinaNet~\cite{yu2025holstein}~\textit{(right)} on detecting cows without fine-tuning: the RetinaNet species detector module from MultiCamCows2024, pretrained on sparsely grouped cows, underperforms in detecting individuals from crowds. The YOLO-v11x model pretrained on the COCO dataset completely fails to detect and localise any cow.}}
\label{fig:obb-vis}
\end{figure}

\ready{Existing architectures~\cite{gao2022label, yu2025holstein} for object detection and re-identification (Re-ID) tasks are often proficient in accurately localising scattered targets. In the field of wildlife monitoring~\cite{vcermak2024wildlifedatasets} and farm animal surveillance~\cite{kaur2023cattle, moosa2024self}, such Re-ID information for species such as zebras, snow leopards, and cows underpins many downstream tasks including animal behaviour monitoring and ecological management. Outline-breaking, high-contrast coat patterns in some of these species~(e.g. zebra stripes, cow patches) can result in significant underperformance in detectors as conspecifics gather closely~(see Figure~\ref{fig:obb-vis}) -- producing a `dazzle' effect~\cite{lovell2024dazzle}. For grouped Holstein-Friesian cows in particular, the performance of standard object detection pipelines such as YOLO~\cite{khanam2024yolov11} breaks down or is significantly impaired due to dazzle-induced complications regarding boundary detection~\cite{lovell2024dazzle} -- typically resulting in poor animal segmentation.}

\begin{figure}[!htbp]
\centering
\captionsetup{justification=centering}
\begin{tabular}{p{0.49\textwidth} p{0.49\textwidth}}
\begin{subfigure}[b]{0.49\textwidth}
    \includegraphics[width=\linewidth]{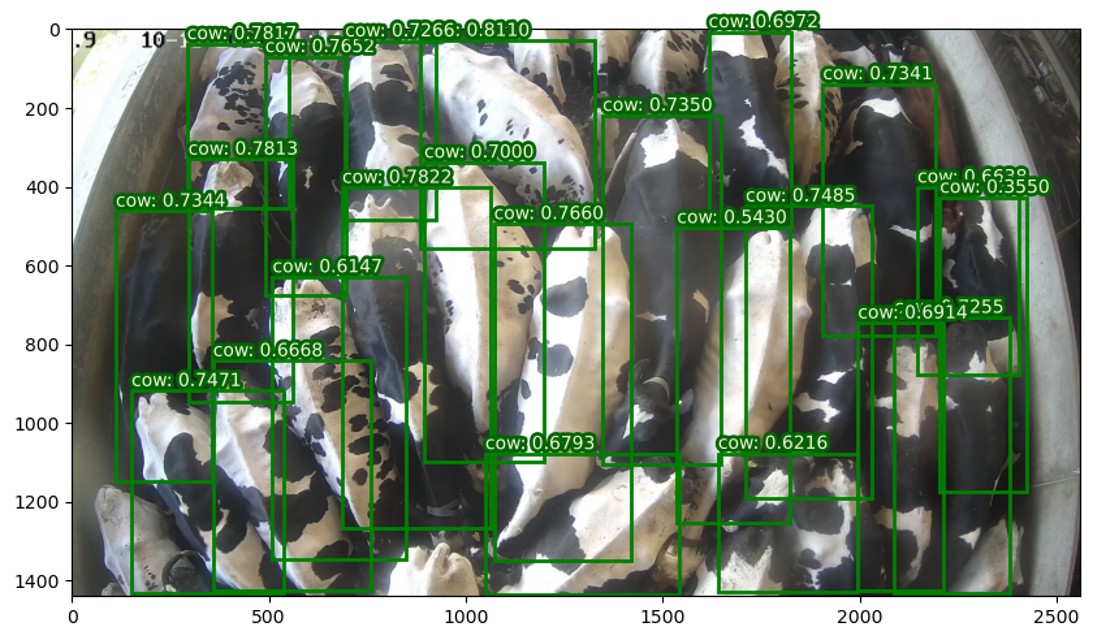}
    \caption{OWLv2}
\end{subfigure} &
\begin{subfigure}[b]{0.49\textwidth}
    \includegraphics[width=\linewidth]{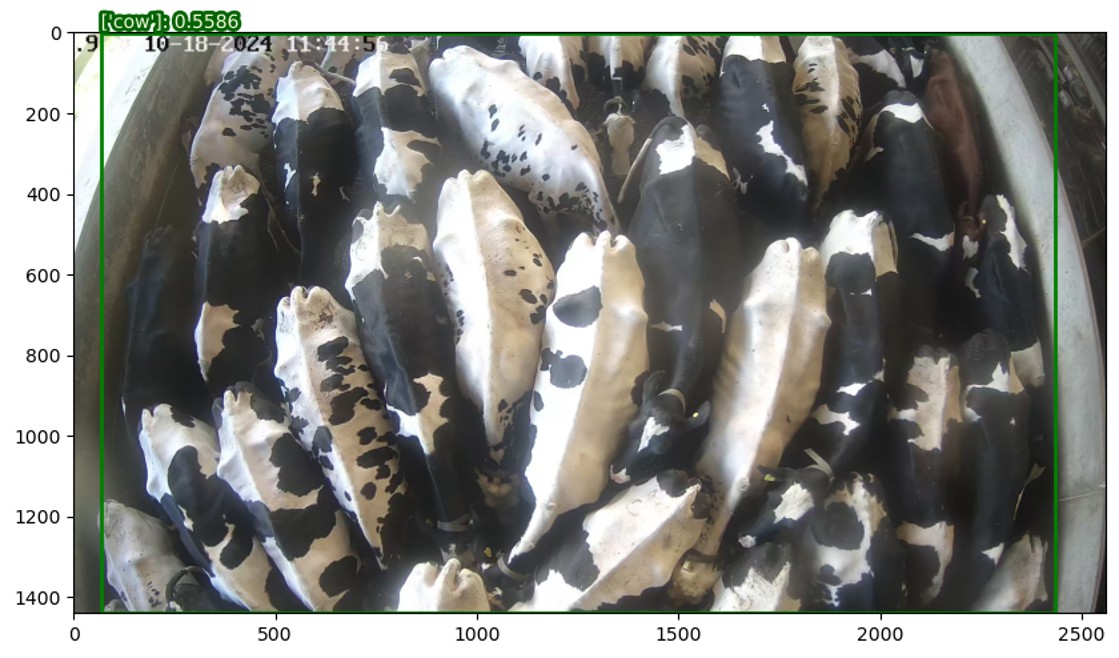}
    \caption{GroundingDINO}
\end{subfigure}
\end{tabular}
\caption{\ready{\textbf{Bounding Box Results from Text-Prompted Detection Models.} The inference results of both OWLv2~\cite{minderer2023scaling} and GroundingDINO~\cite{liu2024grounding} on our cow herds with no prior fine-tuning. GroundingDINO, though much faster than OWLv2 inference, fails to identify individuals when using the same text prompt as input.}}
\label{fig:comp-perf1}
\end{figure}

\begin{figure}[!htbp]
\centering
\captionsetup{justification=centering}
\begin{tabular}{p{0.49\textwidth} p{0.49\textwidth}}
\begin{subfigure}[b]{0.49\textwidth}
    \includegraphics[width=\linewidth]{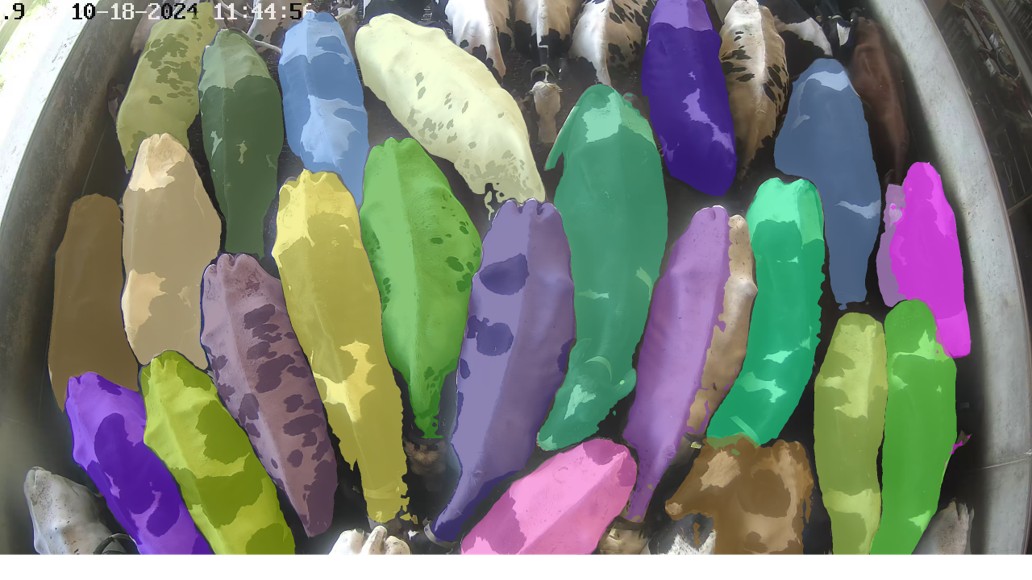}
    \caption{OWL + SAM2 \textbf{(ours)}}
\end{subfigure} &
\begin{subfigure}[b]{0.49\textwidth}
    \includegraphics[width=\linewidth]{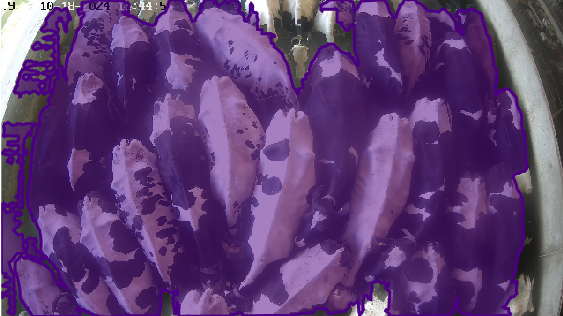}
    \caption{GroundedSAM}
\end{subfigure}
\end{tabular}
\caption{\ready{\textbf{Instance Segmentation on Crowded Holstein-Friesians.} Illustrations of instance segmentations from our pipeline~\textit{(left)} and in failure of segmentation with GroundedSAM~\cite{ren2024grounded} when used without prior fine-tuning~\textit{(right)}. GroundedSAM and our pipeline are, in principle, both capable of segmenting polygon masks based on text inputs. Yet GroundedSAM, segmenting one single region using the same text prompt `cow' as our pipeline, or a varied selection of prompts such as `a cow', `individual cow', or `cows', fails to separate and outline the torso of each cow individually. Other prompts, such as `all instances of cows among a group' or `30 individual cows in the densely packed group', fail to provide any detection for OWLv2, or segmentation for GroundedSAM.}}
\label{fig:comp-perf2}
\end{figure}

\ready{To detect individuals in the presence of such `dazzle' phenomena, pixel-level information is needed to replace CNN-based bounding boxes. The advent of attention-based~\cite{ravi2024sam2}, prompted~\cite{radford2021learning, minderer2023scaling, oquab2023dinov2} and depth~\cite{yang2024depth, yang2024depth2} visual models has introduced a family of pixel-level operating frameworks with reduced latency and increased performance. They detect and localise more precisely by maintaining complex morphologies of targets compared with bounding boxes obtained from standard convolution~\cite{dosovitskiy2020image}. Once trained, these models offer high levels of performance with little or no additional input. The ability to segment via text prompts, combined with their capability to embed visual features alongside language, gives rise to faster annotation models~\cite{jiang2025detect}. During ground-truth labelling, rather than an exhaustive and error-prone manual drawing of the bounding boxes, the prompted models use text describing the targets to be detected. Text-prompted, or grounded, models such as GroundingDINO~\cite{liu2024grounding} and GroundedSAM~\cite{ren2024grounded} further leverage the visual features of trained models for segmenting targets. However, they still suffer from under-segmentation when detecting individuals in a group of densely located Holstein-Friesian cows, as shown in Figures~\ref{fig:comp-perf1} and~\ref{fig:comp-perf2}.}

\ready{Utilising large grounded models, we propose a Re-ID pipeline that combines existing Open-Vocabulary Weight-free Localisation~(OWLv2)~\cite{minderer2023scaling} with the Segment Anything Model~(SAM2)~\cite{kirillov2023segment} to automate cow identification of closely grouped Holstein-Friesian cattle. We first acquire the data from a two-step OWLv2-SAM2 pipeline from farm footage. Then, as a comparative analysis, we compare the segmentations with standard YOLO, RetinaNet and baseline SAM2 models for this task versus a human-derived ground-truth. After data acquisition, the segmentations are used to demonstrate individual (re-)identification capability. Using an unsupervised contrastive learning~(UCL) framework, we test the identification performance for these individual cows based on multiple days of indoor farm footage. The overall pipeline is visualised in Figure~\ref{fig:overview}. The code for the pipeline and the dataset~\footnote{\href{https://phoenix4582.github.io/dazzlecows.github.io/}{https://phoenix4582.github.io/dazzlecows.github.io/}}are available for reproducibility.}

\section{Background}
\label{sec:background}

\subsection{Re-Identification for Animals}
\label{sec:subsec02-2-1}

\ready{The Re-Identification task is one of the most well-known challenges for performance and transferability evaluation of the object segmentation models. The task aims to detect known targets that reappear with different poses, appearances, locations, or backgrounds~\cite{ye2021deep, ye2025transformer}. This pipeline has a wide application, not solely restricted to wildlife~\cite{vcermak2024wildlifedatasets, adam2025wildlifereid}, pedestrian~\cite{behera2023large} and vehicle~\cite{qian2023urrnet} monitoring.}

\textbf{Surveillance of Farm Animals.} \ready{Farm animal surveillance is an important application of the Re-ID pipeline application. Gao et al.~\cite{gao2021towards, gao2022label} constructed a simple, yet effective, self-supervision pipeline for the Re-ID of indoor Holstein-Friesian cows based on RetinaNet and contrastive learning. Wang et al.~\cite{wang2022towards} used a one-shot tracker for both the detection and Re-ID of housed pigs for long-term monitoring. Implemented on a variety of coat-pattern themed datasets, containing a range of occlusion, brightness and camera angles~\cite{bhole2019computer, gao2021towards}, Dubourvieux et al.~\cite{dubourvieux2023cumulative} introduced a Cumulative Unsupervised Multi-Domain Adaptation~(CUMDA) strategy for Re-ID over Holstein-Friesian cows to monitor health and body condition.}

\textbf{Data Acquisition Bottleneck.} \ready{We note that all aforementioned Re-ID pipelines and most approaches in this general field require an arduous data-set acquisition phase. In our case---crowded cow settings---biological patterns and formations pose a challenge when manually annotating thousands of segmentations. This bottleneck in dataset preparation also brings reduced accuracy when the Re-ID pipeline is transferred for use on different farms. By inheriting knowledge from general large-scale models, built from huge sets of data (\cite{deng2009imagenet, lin2014microsoft}), we are able to successfully identify Holstein-Friesian cows with reduced manual labelling and increased transferability.}

\subsection{Prompt-based Object Detection}
\label{sec:subsec02-2-2}

\textbf{Overview.} \ready{Recently developed prompt-based object detection models can be utilised to effectively detect individuals from grouped Holstein-Friesian cows due to these models' adaptability, reduced latency and pixel-level precision~\cite{jiang2024t, ravi2024sam2}. Object detection pipelines in surveillance systems require sufficient pre-labelled data to train the model so as to identify and localise targets. For precise object detection tasks---both salient~\cite{wang2021salient} and camouflaged~\cite{xiao2024survey}---pixel-level segmentation replaces bounding boxes for detailed representation. However, manually preparing fully labelled datasets covering all possible object variations suffers from a longer preparation period and low transferability~\cite{fouquet2023transferability}. The preparation of bounding boxes or pixel-level segmentations benefits from automation.}

\textbf{Text-Vision Integration.} \ready{By using simple text inputs, autonomous labelling of data using Contrastive Language-Image Pre-training~(CLIP)~\cite{radford2021learning} introduced the potential for easily creating the labelled data necessary for training large models. They embedded both text prompts and the corresponding visual input to learn correlations, such as animal species and localisation. This approach enables a simplified data generation process, since it can be used with little human intervention on new inputs. The work demonstrated its adaptability and transferability on various tasks, such as Optical Character Recognition, action-recognition in videos~\cite{liu2025cattle}, geo-localisation and Fine-Grained Image Analysis.}

\textbf{Text-Visual Abstraction.} \ready{Enhancements to CLIP~\cite{xu2023learning} have shown that object localisation can also be chained with text prompts to allow automated data preparation for object detection. The work demonstrated pixel-level feature learning over a weakly supervised large set of objects, providing accurate localisation. In addition to a global representation of an entire image, their model embedded target localisation along with the text inputs. Following an efficient error-checking phase, datasets of localised targets were acquired rapidly in stages.}

\ready{Open-world Localisation~\cite{minderer2022simple} and Grounding DINO~\cite{liu2024grounding} were further developed with higher volumes of text and visual input, leading to increased segmentation precision. They also constitute an essential part of some of the most widely used multi-modal models~\cite{ye2023mplug, chen2024plug}. Extending the earlier version of simple text-visual feature pairing from CLIP that used a limited vocabulary, OWL pairs free-form queries with visual features to learn text-visual correlations. The model is capable of understanding implicitly associated information within text prompts~(e.g., properties of entities mentioned) to target objects with complex definitions. Grounding DINO further expands the utility of text encoders to inject contrastive feedback into multiple parts of visual encoders, leading to open-set capabilities.}

\textbf{Maximising Feature Utility.} \ready{The Segment Anything Model~(SAM)~\cite{kirillov2023segment} exploits vision features more deeply by introducing a greater variety of visual inputs as prompts. The model allows the detailed segmentation of images using highlight points, bounding boxes and polygon masks. Following the pattern of query-key pairing inherent in the Transformer pipeline, they allow multi-staged pairing for learning features and representations. Grounded SAM~\cite{ren2024grounded} merges Grounding DINO's ability to understand text with SAM to guide segmentation. State-of-the-art medical instance segmentation~\cite{zhang2023customized, zhu2024medical, wu2025medical} and image processing~\cite{zhang2023personalize, yu2023inpaint, liu2023any} applications have demonstrated the application of SAM-based approaches as a promising successor to traditional, labour-reliant, counterparts based on CNNs.}

\subsection{Unsupervised Contrastive Learning}
\label{sec:subsec02-2-3}

\ready{Contrastive learning has emerged as a powerful approach for representation learning, often employed in self-supervised settings for Re-ID tasks. However, the field of unsupervised contrastive learning, which operates without ground-truth throughout the learning phase, remains an important and less-explored frontier. To build smarter and more competitive farm animal surveillance systems, unsupervised approaches promise improved transferability without costly labelling.}

\textbf{Theoretical Foundations.} \ready{Unsupervised contrastive learning aims to meaningfully extract representations by projecting images into a space of reduced dimensionality, maximising the similarity between intra-class instances and maximising inter-class separation. Unlike self-supervised contrastive learning, which relies on labels, unsupervised methods generally explore intrinsic data structures and statistical correlations instead. Several key concepts define this approach, including instance discrimination, clustering-based objectives and mutual information maximisation. These enable clustering by assigning samples as groups without explicit labels. Learning to preserve mutual information across transformations is one of the main challenges for the unsupervised pipeline.}

\textbf{Practical Applications.} \ready{One of the first unsupervised contrastive learning (UCL) methods treated each data sample as a unique class. The method proposed by Wu et al.~\cite{wu2018unsupervised} introduced a nonparametric softmax classifier that learnt representations by distinguishing individual instances without explicit labels. Approaches such as Deep InfoMax by Hjelm et al.~\cite{hjelm2018learning} leverage mutual information estimations to learn high-level representations without explicit supervision. Between the boundaries of self-supervised and unsupervised learning, methods involving SwAV by Caron et al.~\cite{caron2020unsupervised} replace explicit contrastive pairs with clustering assignments, dynamically grouping similar data points whilst maintaining discriminative features. In 2023, Zang et al.~\cite{zang2023diffaug} introduced DiffAug using domain-specific augmentations for contrastive learning in an unsupervised setting. Prior, Contrastive Multiview Coding(CMC)~\cite{tian2020contrastive} learned from multiple perspectives of data to enhance representation learning. In 2024, the implementation of CLMIA~\cite{chen2024clmia} utilised contrastive learning for membership inference attacks.}

\textbf{Challenges.} \ready{Common issues for UCL include learning with no negative examples, evaluating models without downstream supervision, and handling data heterogeneity. For our cow Re-ID task, each cow appears only once per timestamp. Therefore, in this specific case, each cow becomes a separate class. These images can be easily augmented to improve performance, helping to solve issues around the imbalanced appearance of certain cows.}

\section{Methodology}
\label{sec:methodology}
\subsection{Automated Mask Extraction}
\begin{figure}[!ht]
\centering
\captionsetup{justification=centering}
\includegraphics[width=12.5cm]{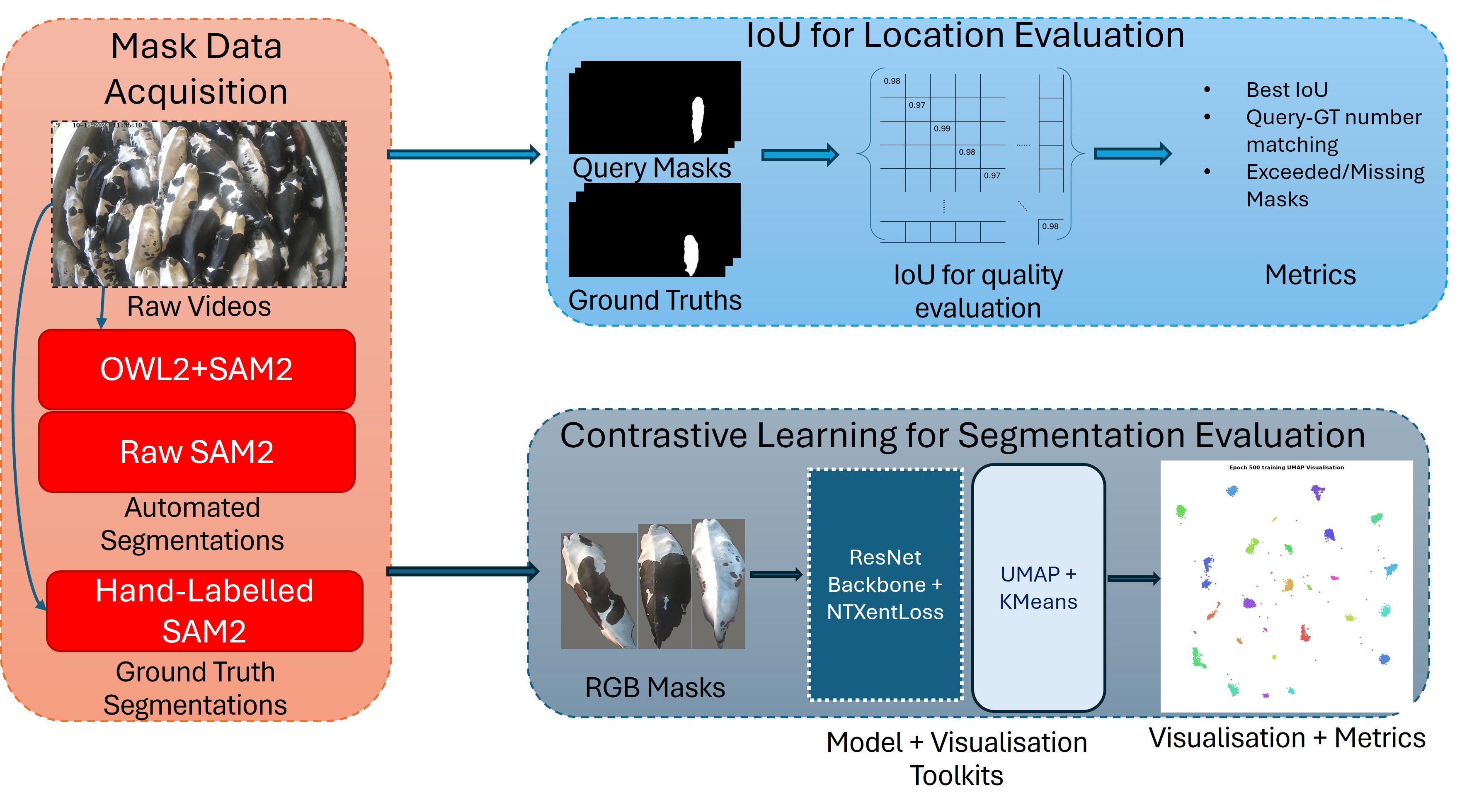}
\caption{\ready{\textbf{Processing Pipeline Overview.} Our framework consists of two main components. For data acquisition~\textit{(left)} of both mask segmentations and ID, we initiate tracking and relocate targets at $1s$ intervals via OWLv2-enhanced and baseline SAM2 models. For ground truth image set generation, we manually label the bounding boxes and use them as the initial input for tracking. Then, two sections of evaluation were performed. We first apply IoU matching between SAM2 binary masks and ground truth to evaluate their localisation performance~\textit{(upper-right)}. After integrating binary masks with their original frames to get the RGB masks, we applied a UCL module to evaluate Re-ID performance~\textit{(lower-right)}.}}
\label{fig:overview}
\end{figure}

 \ready{Our pipeline begins with the acquisition of Holstein-Friesian image data without human intervention. Our two-stage approach shown in~Figure~\ref{fig:overview} incorporates OWLv2 with the SAM2 model to extract each cow and streamline autonomous instance segmentation.}
 

\begin{figure}[!ht]
\centering
\captionsetup{justification=centering}
\includegraphics[width=12.5cm]{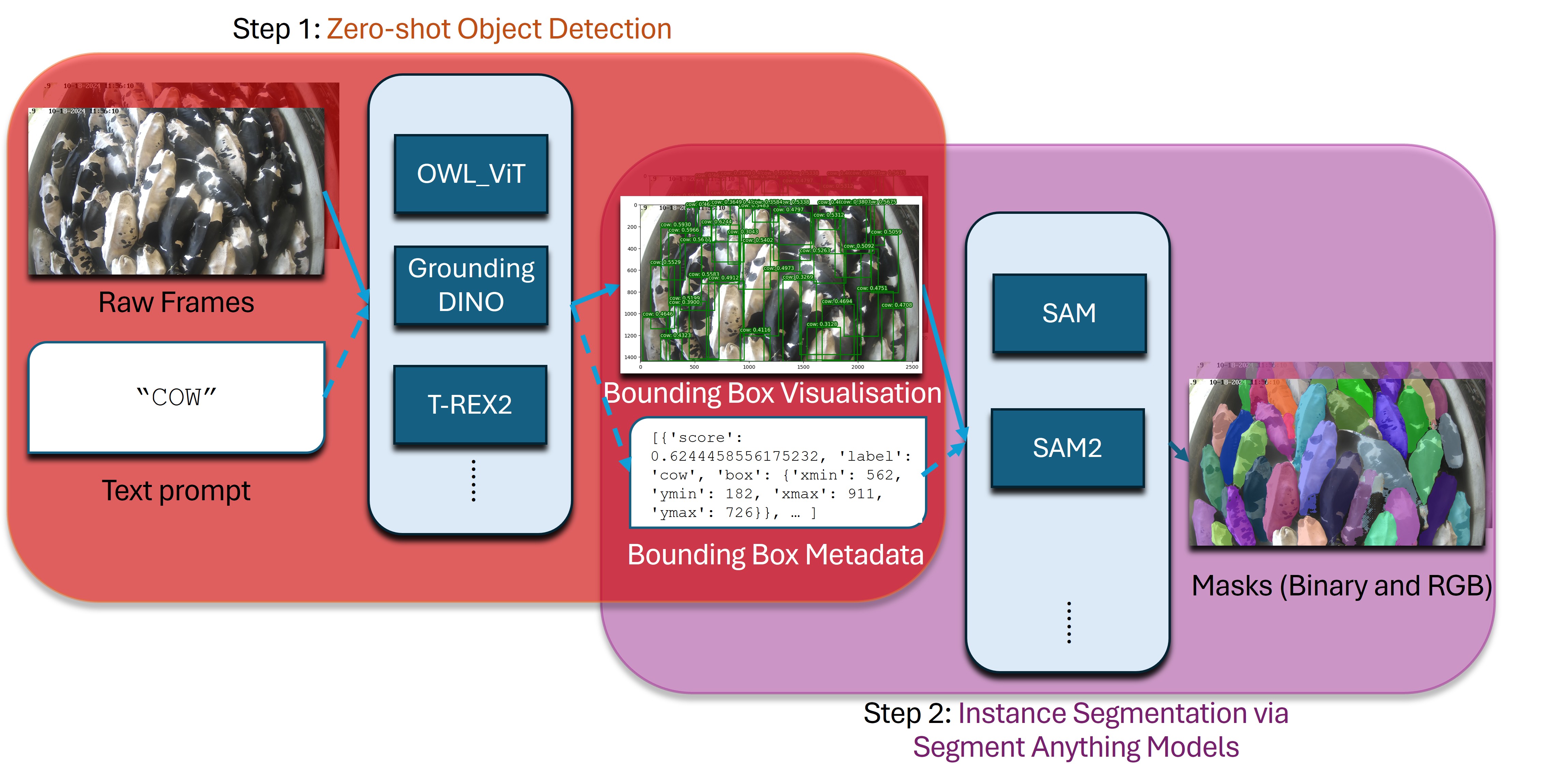}
\caption{\ready{\textbf{Automated Data Acquisition.} Our proposed approach for using prompts to generate segmentation masks has two stages: first in~\textit{(red)}, we use a text prompt along with video frames as the inputs to a pre-trained OWLv2~\cite{minderer2023scaling} model to get axis-aligned bounding boxes. Then, in~\textit{(purple)}, the same video frames along with the bounding box outputs are used as inputs to a pre-trained SAM2~\cite{kirillov2023segment} model to yield segmentations.}}
\label{fig:overview_data}
\end{figure}

\textbf{Bounding Box Generation.} \ready{As demonstrated in Figure~\ref{fig:overview_data}, we first deploy the OWLv2 model with text prompts alongside video frames to extract bounding boxes using standard weights\footnote{https://huggingface.co/google/owlv2-large-patch14-ensemble} with the single word prompt~`cow'. Then, we refine the bounding boxes through heuristic-based filtering to remove unwanted duplicate segmentations and noise. Specifically, an aspect ratio filter between $2.5\%$ to $7.5\%$ of original frames is used to remove bounding boxes that were either oversized or undersized. Then, non-maximum suppression filtering is used to remove duplicates.}

\textbf{Mask Acquisition.} \ready {After gathering the refined bounding boxes, we feed these inputs into the Ultralytics pre-trained SAM2-2.1L model\footnote{https://github.com/ultralytics/assets/releases/download/v8.3.0} to extract binary masks and track each cow from frame to frame. Masks are generated from  SAM2 using the previous OWL bounding boxes as inputs once per second.}

\textbf{Species Detector Evaluation.} \ready{We compare the performance of this approach versus a standard SAM2 baseline model -- the only input being images. Both our results and the baseline outputs are compared with manually annotated ground truth masks.}

\subsection{Evaluation via Unsupervised Contrastive Learning}

\begin{figure}[!ht]
\centering
\captionsetup{justification=centering}
\includegraphics[width=12.5cm]{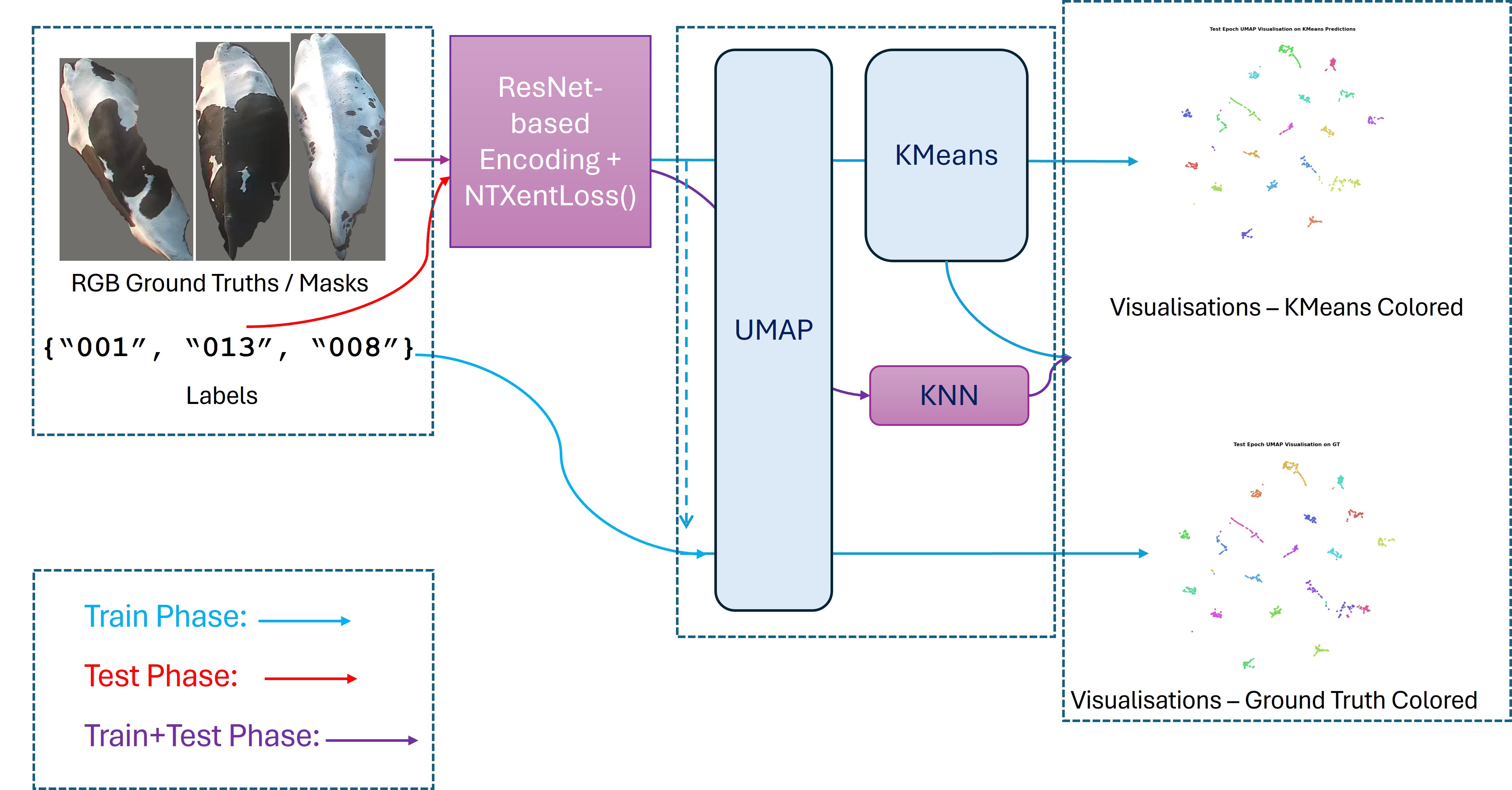}
\caption{\ready{\textbf{Unsupervised Contrastive Learning Module Overview.} During the training phase, we feed the masks from different individuals within the same time instance into our NTXentLoss function. The learned embeddings are analysed using K-Nearest Neighbours, and K-Means is used for clustering. UMAP is used for visualisations of raw embeddings coloured with the ground-truth labels, and also with K-Means labels for inference.}}
\label{fig:overview_ucl}
\end{figure}

\textbf{Individual Cow Reconstruction.} \ready{Once binary masks are acquired, the next step is to utilise these for individual cow Re-ID. We have converted the binary masks with the corresponding video frame into RGB masks to extract cow skin patterns in a way that considers morphology information. We have used the DC component of the frame as the background of segmented frames to minimise foreground-background discrepancy. The resulting dataset consists of RGB masks for each cow from every sampled timestamp.}

\textbf{Embedding the Features.} \ready{With ID-encoded RGB masks as input, we then fine-tune a ResNet\cite{he2016deep} model to embed the masks for contrastive learning~(see Figure~\ref{fig:overview_ucl}). We combined a timestamp-instanced sampling strategy with the NTXentLoss~\cite{chen2020simple}. Given that all masks sampled from the same timestamp were guaranteed to represent different individuals, we traverse only one specific timestamp every training epoch to avoid giving the same cow different labels. During both validation and testing, we use k-Nearest Neighbour~(kNN) clustering to analyse whether the model is able to classify the data effectively. We also use the Adjusted Random Index~(ARI)~\cite{hubert1985comparing}, Adjusted Mutual Information~(AMI)~\cite{vinh2009information}, Normalised Mutual Information~(NMI) \cite{amelio2017correction}and accuracy of the Hungarian Algorithm~(HA)~\cite{kuhn1955hungarian} on the predictions for a statistical understanding of clustering performance.}

\subsection{Implementation of Re-Identification}
\begin{figure}[!htbp]
\centering
\captionsetup{justification=centering}
\begin{tabular}{p{0.32\textwidth} p{0.32\textwidth} p{0.32\textwidth}}
\begin{subfigure}[b]{0.32\textwidth}
    \includegraphics[width=\linewidth]{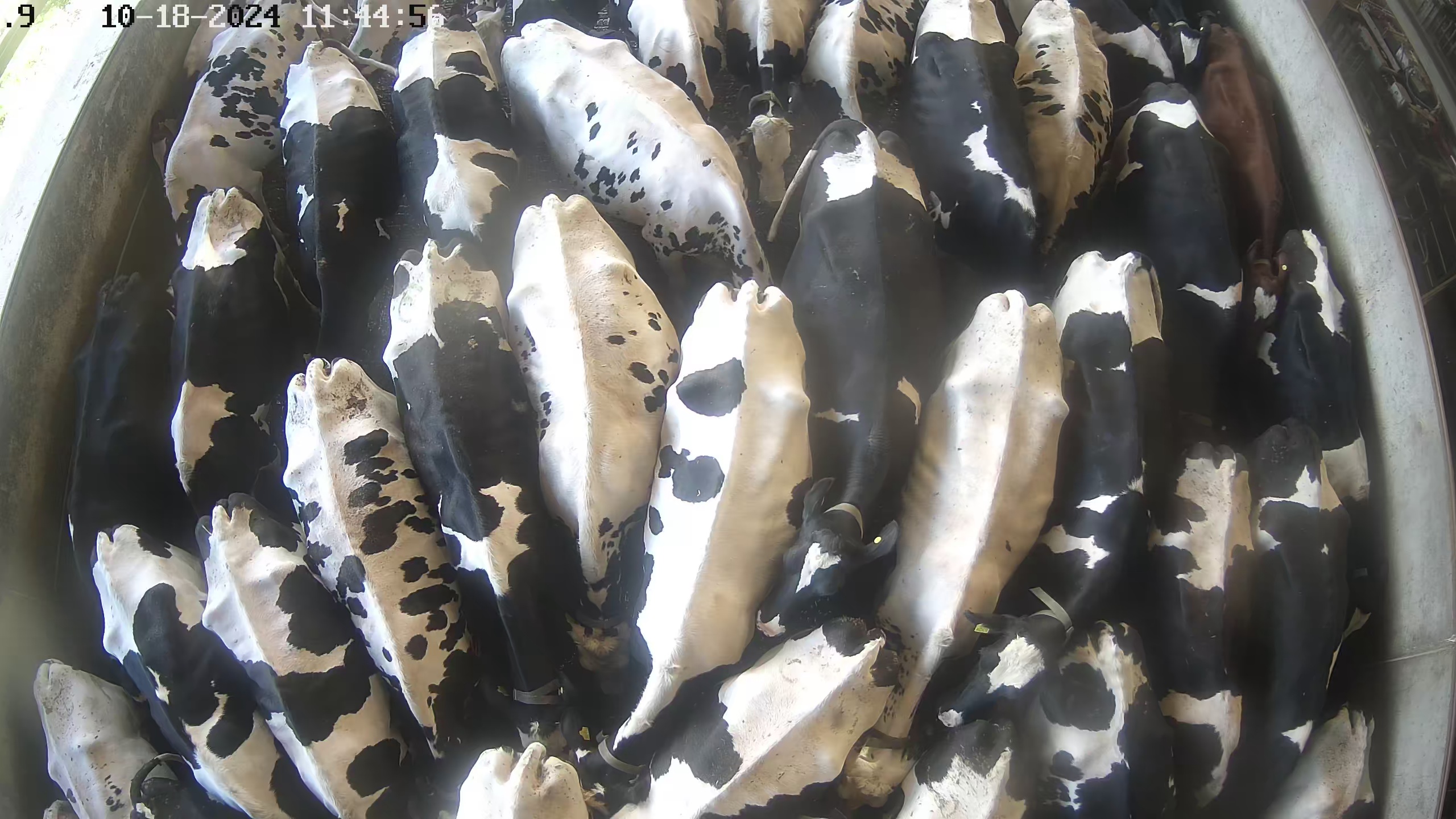}
    \caption{2024-10-18}
\end{subfigure} &
\begin{subfigure}[b]{0.32\textwidth}
    \includegraphics[width=\linewidth]{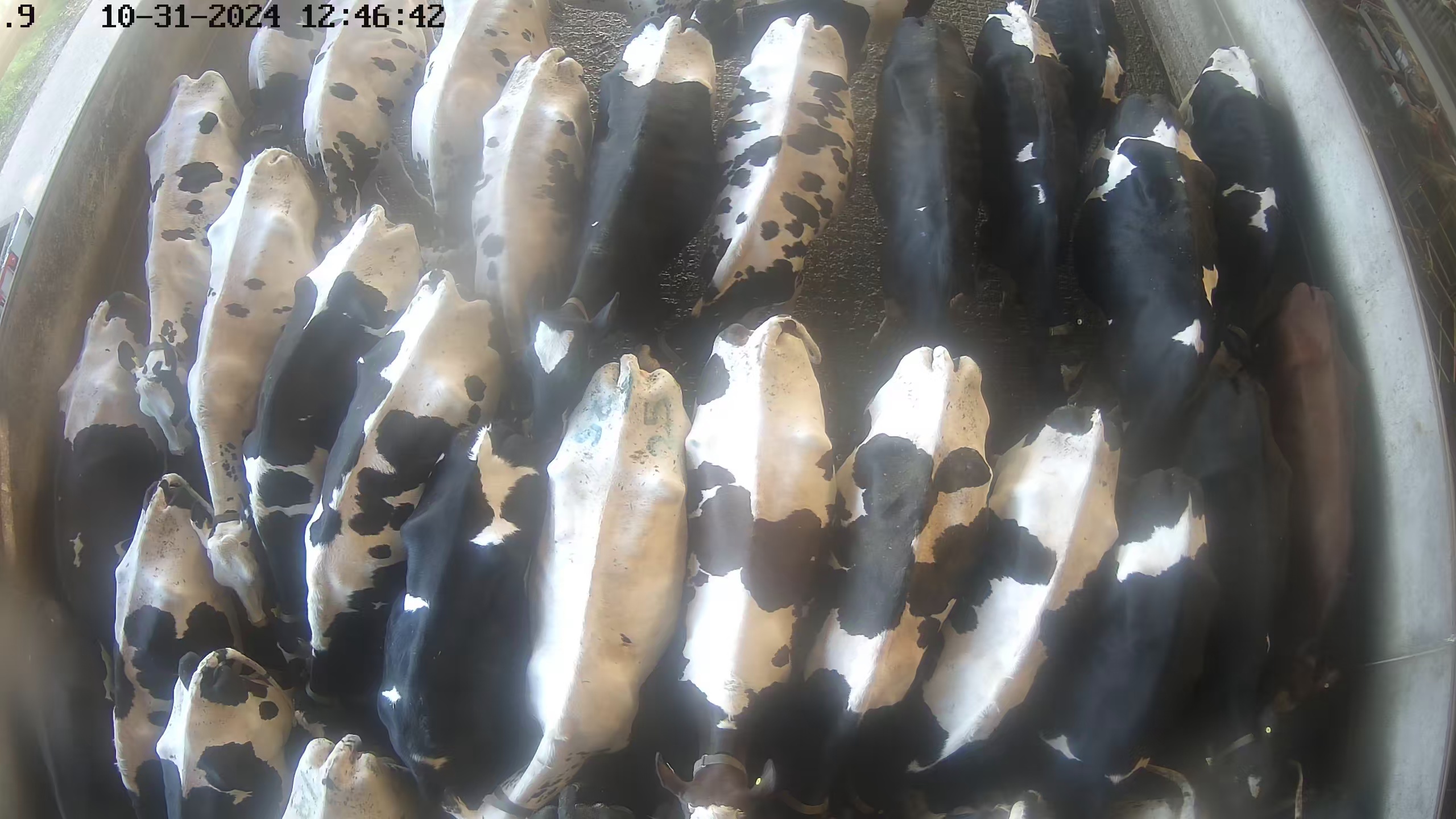}
    \caption{2024-10-31}
\end{subfigure} &
\begin{subfigure}[b]{0.32\textwidth}
    \includegraphics[width=\linewidth]{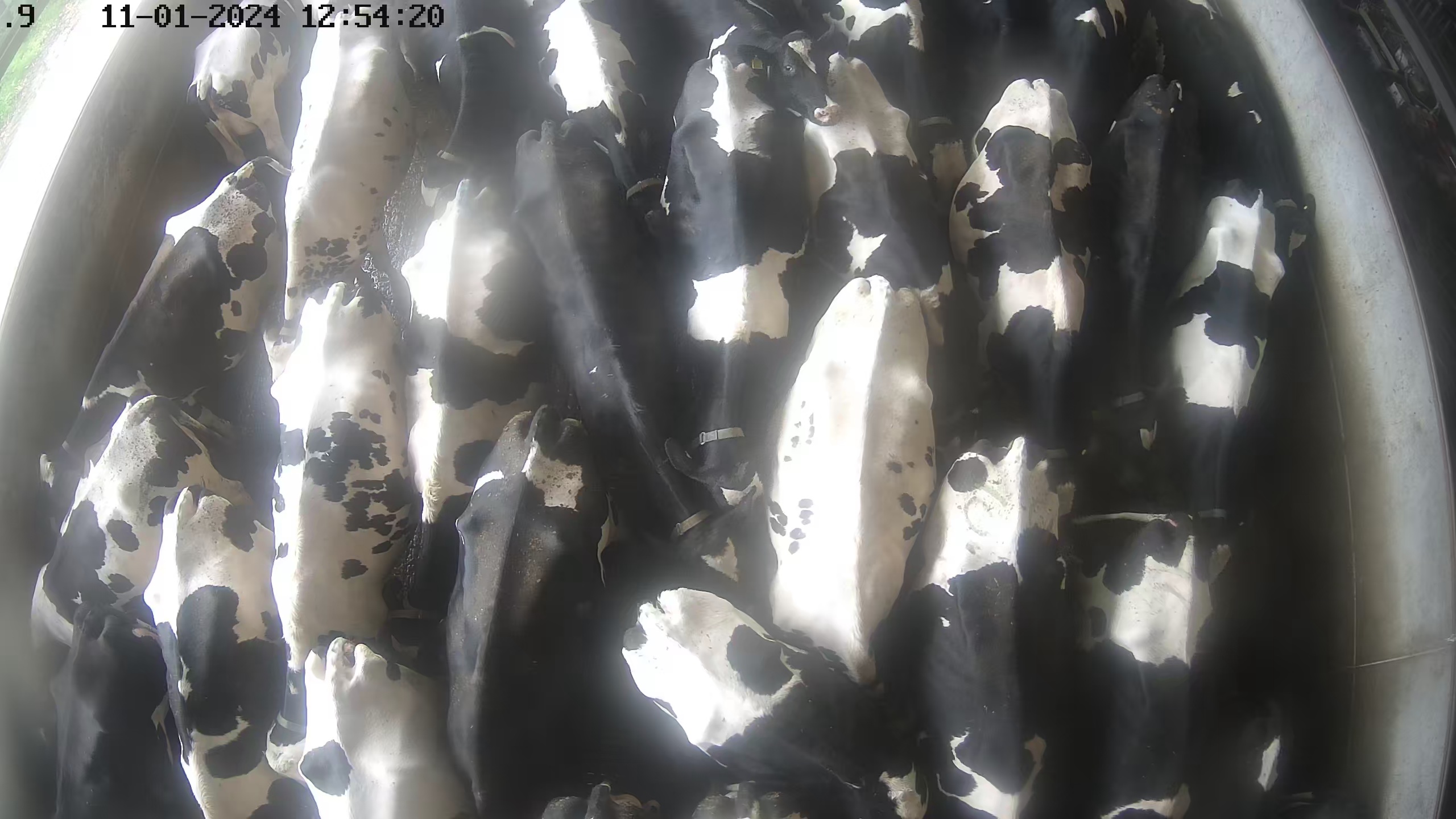}
    \caption{2024-11-01}
\end{subfigure} \\

\begin{subfigure}[b]{0.32\textwidth}
    \includegraphics[width=\linewidth]{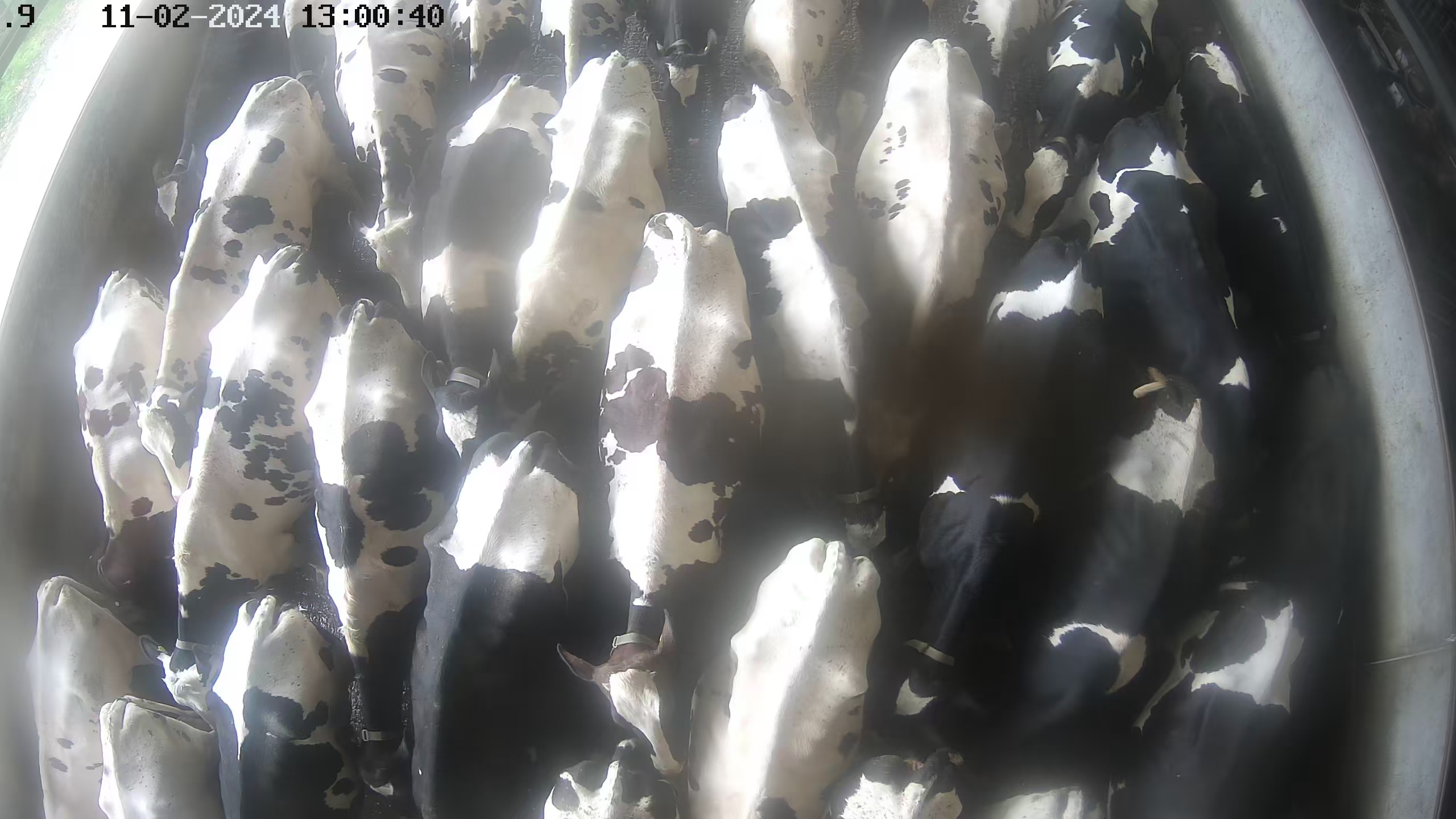}
    \caption{2024-11-02}
\end{subfigure} &
\begin{subfigure}[b]{0.32\textwidth}
    \includegraphics[width=\linewidth]{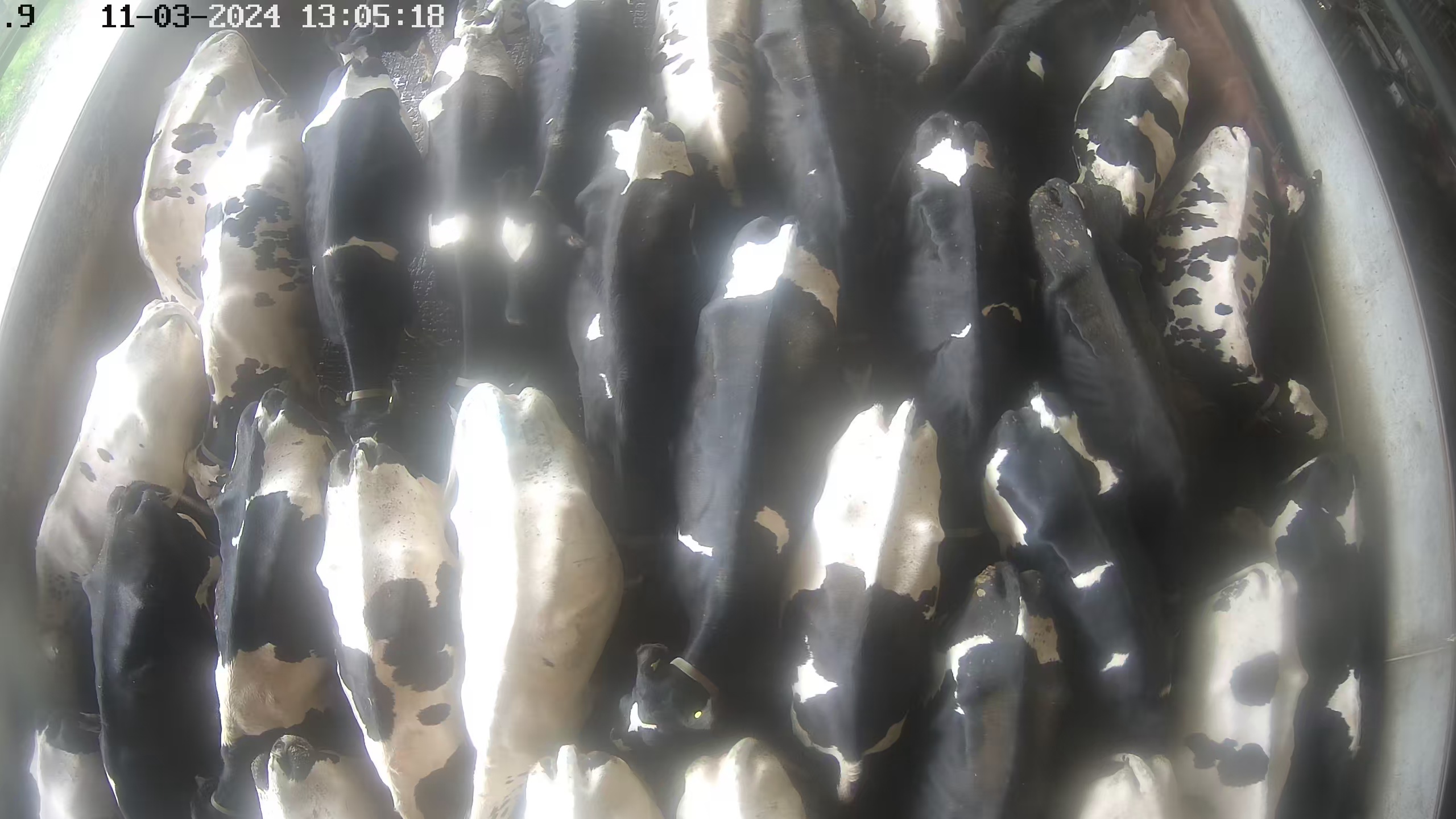}
    \caption{2024-11-03}
\end{subfigure} &
\begin{subfigure}[b]{0.32\textwidth}
    \includegraphics[width=\linewidth]{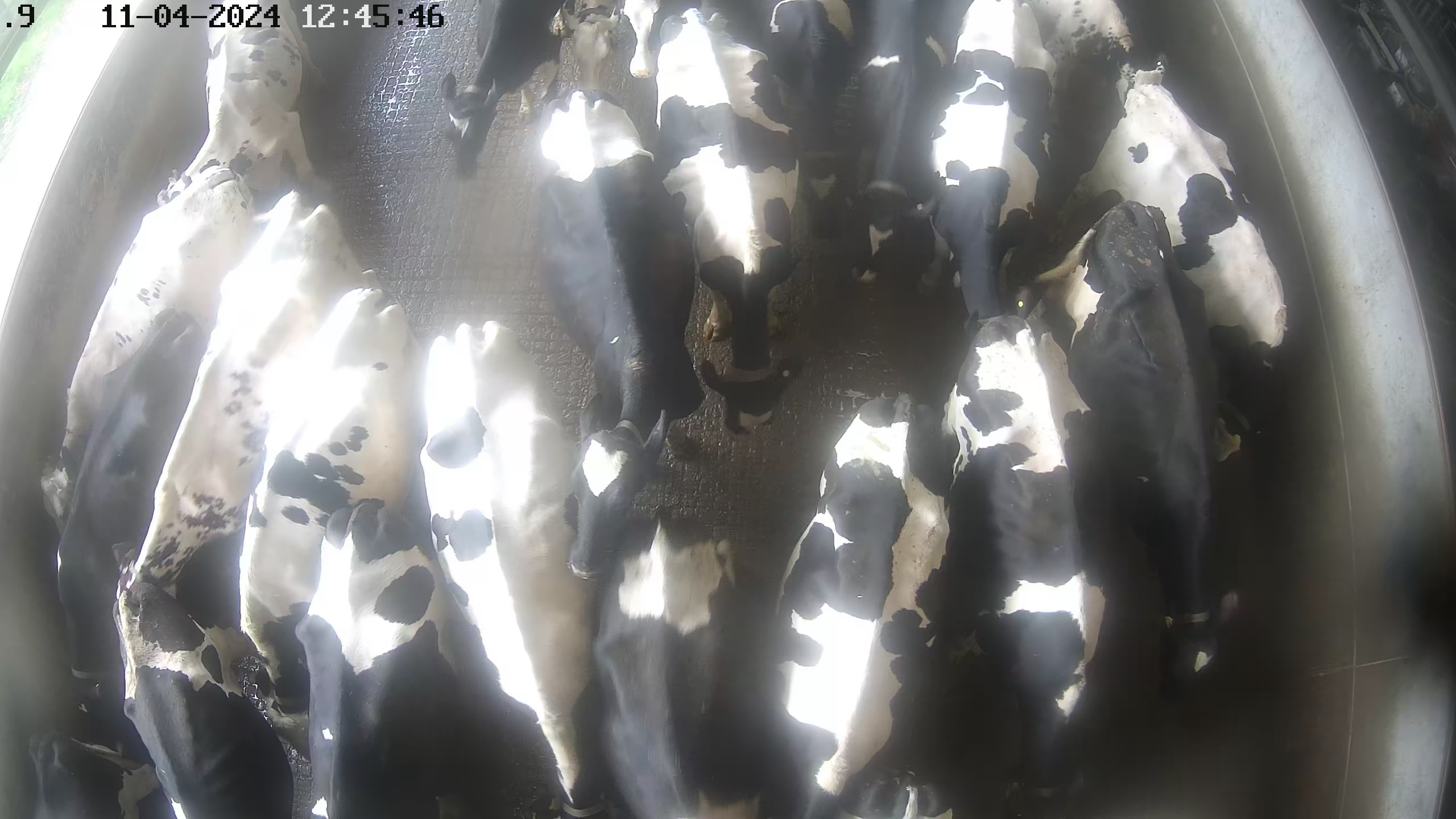}
    \caption{2024-11-04}
\end{subfigure} \\

\begin{subfigure}[b]{0.32\textwidth}
    \includegraphics[width=\linewidth]{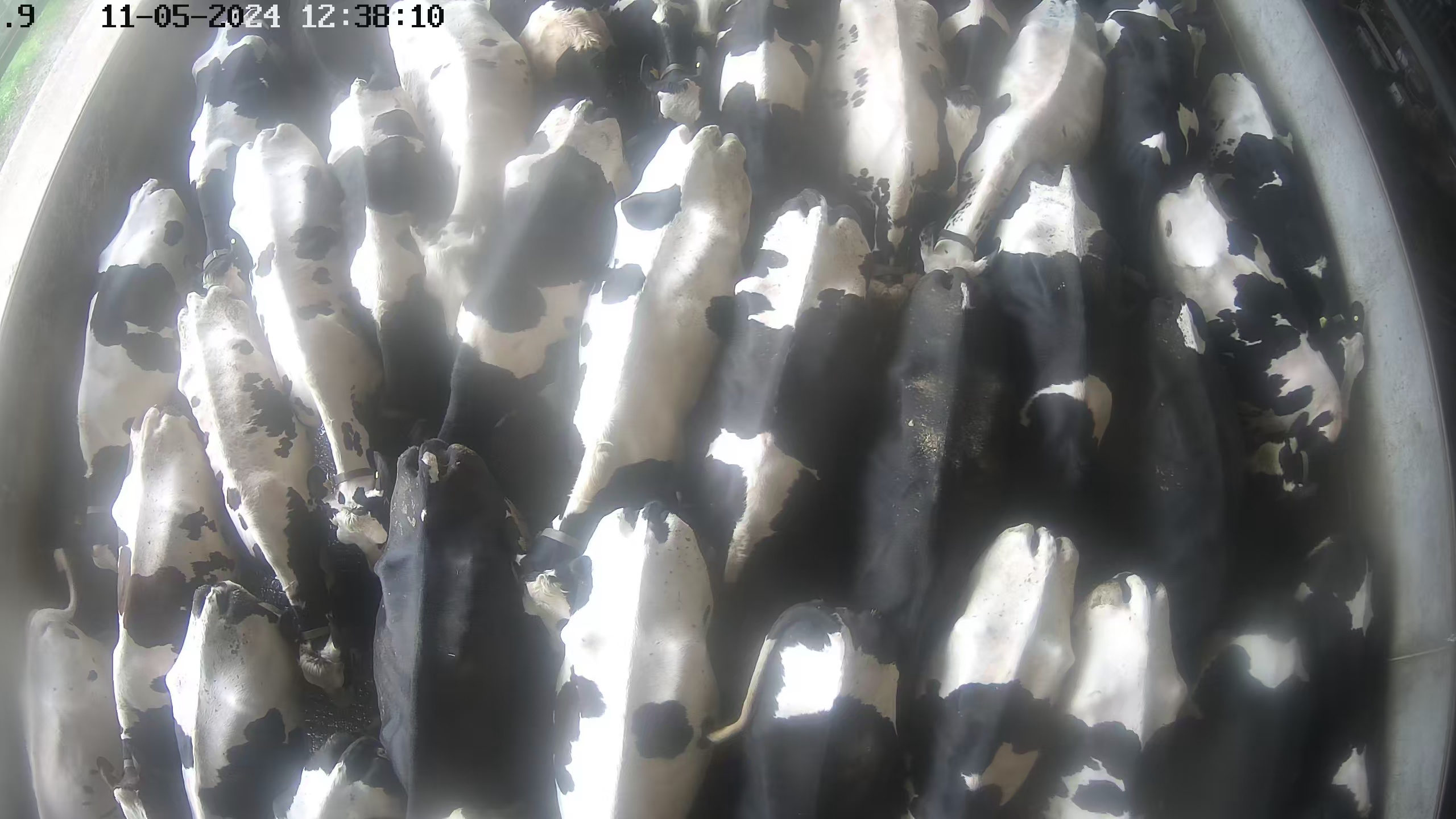}
    \caption{2024-11-05}
\end{subfigure} &
\begin{subfigure}[b]{0.32\textwidth}
    \includegraphics[width=\linewidth]{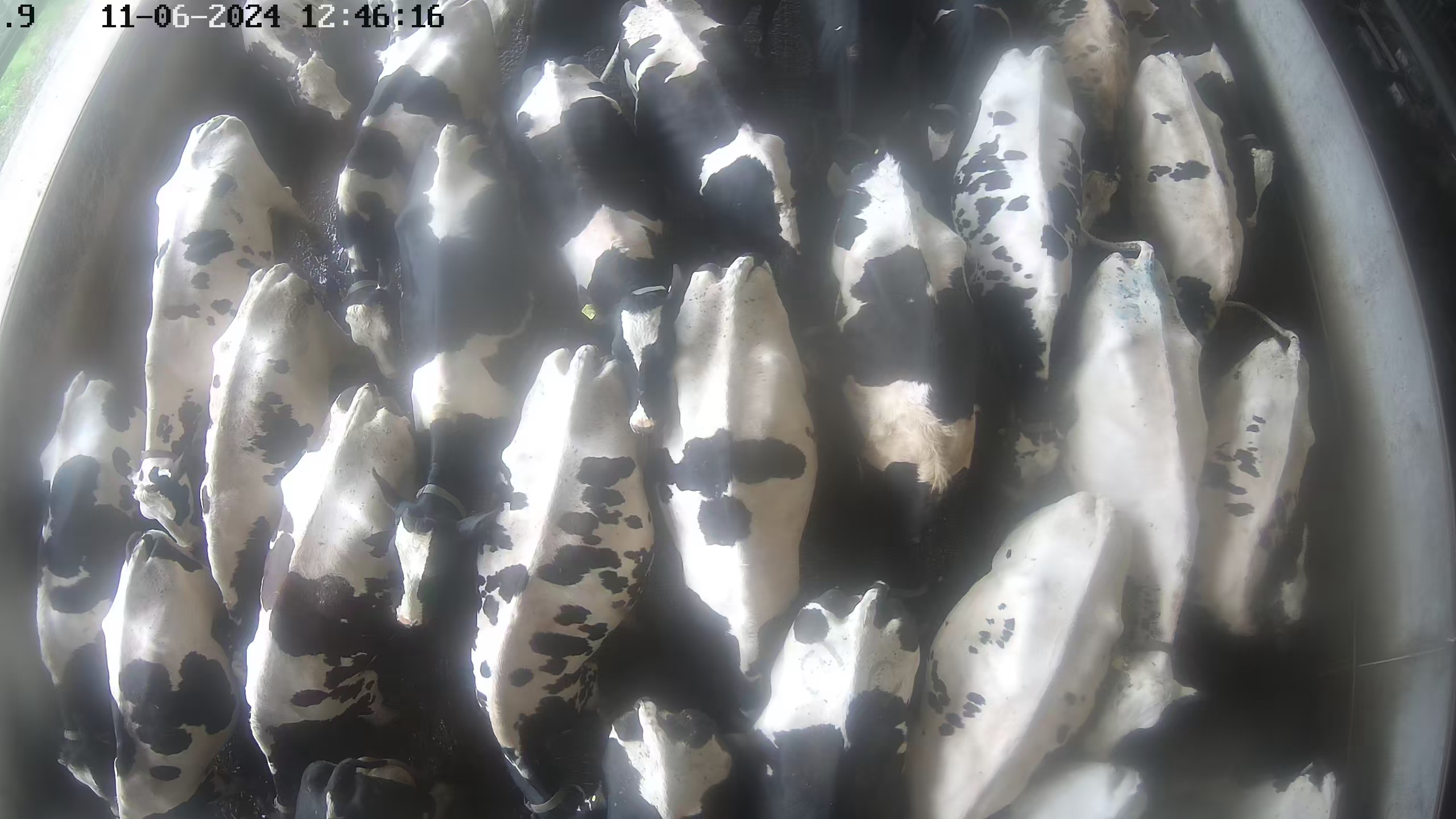}
    \caption{2024-11-06}
\end{subfigure} &
\begin{subfigure}[b]{0.32\textwidth}
    \includegraphics[width=\linewidth]{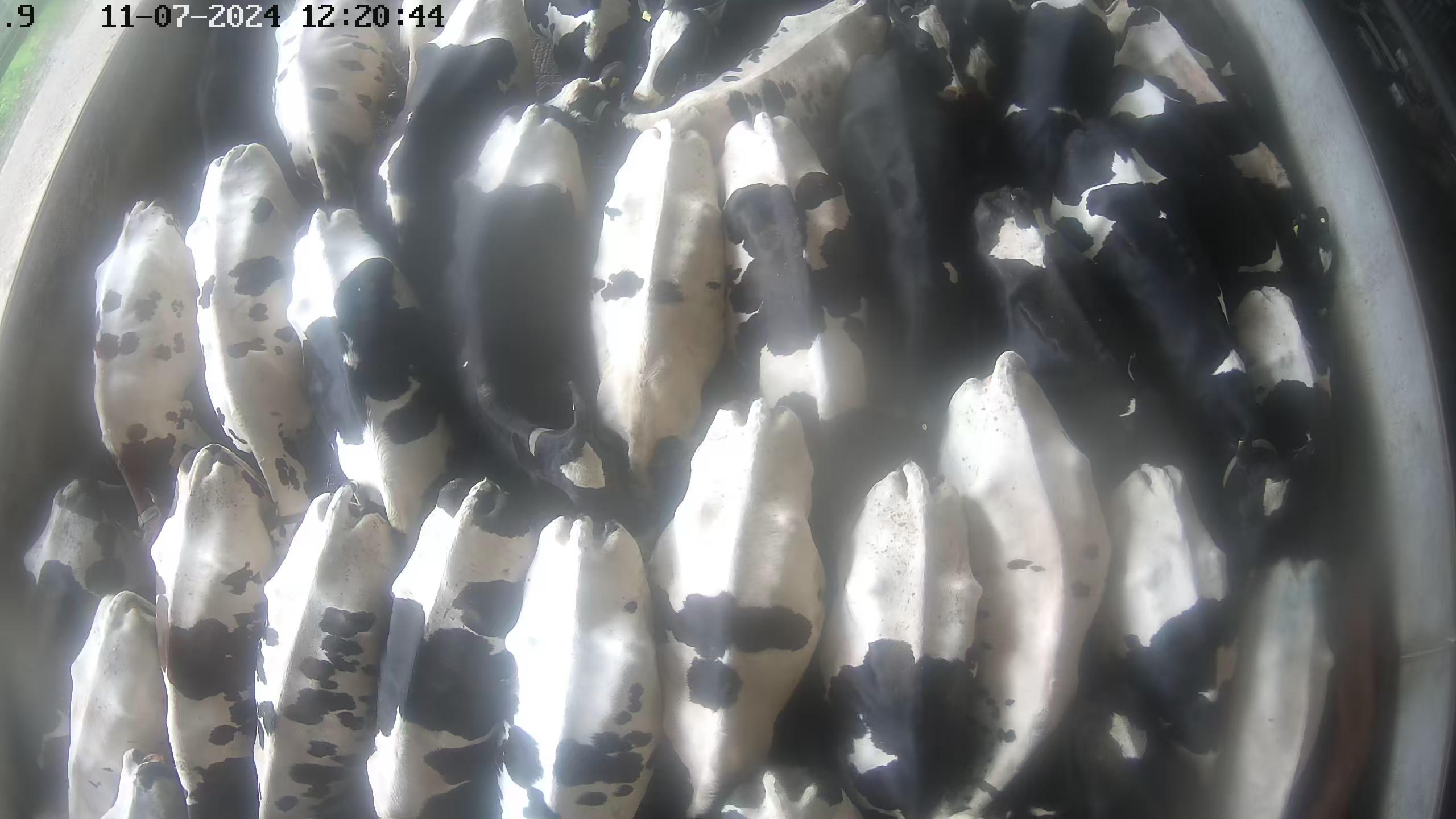}
    \caption{2024-11-07}
\end{subfigure} 

\end{tabular}
\caption{\ready{\textbf{Sample Frames from the Nine Days of the Re-ID Experiment.} Images show a sample frame for each day considered for the re-identification experiment. Note that the timestamp for the first image is one hour off due to summer time adjustment. In addition, as cameras are motion-triggered for recording, the exact time of first capture (after 12:00 pm) varies slightly.}}

\label{fig:daily-overview-9-vis}
\end{figure}

\textbf{Re-ID across Nine Days of Data.} \ready{After obtaining an unsupervised contrastive learner with kNN, we retrained the model from scratch with multiple days of data and evaluated the Re-ID performance. We used a total of $524,469$ RGB masks automatically extracted from nine days of recording in the farm. Using a customised k-fold cross-validation over the nine folds representing a day each, we trained on multiple combinations of seven days' worth of data, leaving unseen folds for validation and testing, respectively. All images for the Re-ID experiment are collected by one camera across the same time window~(12:00pm - 2:00pm) each of the nine days, with individual cows having at most one appearance per day representing one milking of the herd. For those cows that do not attend milking during the reported window no recording is used for this day. Only frames with cows as flagged by OWLv2-SAM2 are used for individual re-identification. The number of cows captured per day varies between 19 to 24 for the nine-day dataset. Figure~\ref{fig:daily-overview-9-vis} shows a sample frame from each of the nine days of the dataset.}

\section{Experiments \& Results}
\label{sec:results}
\subsection{Localisation Evaluation}

\begin{figure}[!htbp]
\centering
\captionsetup{justification=centering}
\begin{tabular}{p{0.49\textwidth} p{0.49\textwidth}}
\begin{subfigure}[b]{0.49\textwidth}
    \includegraphics[width=\linewidth]{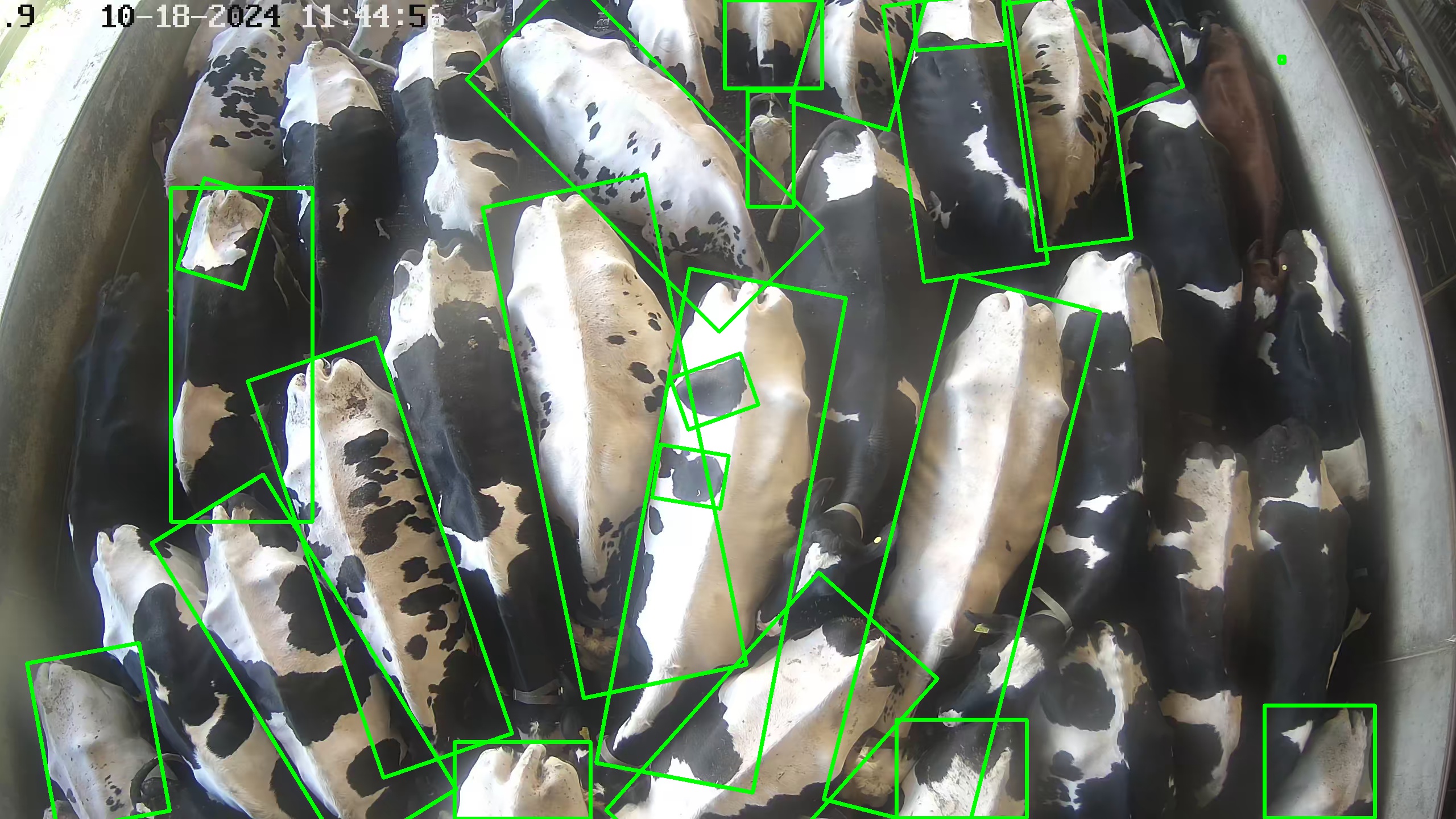}
    \caption{Standard SAM Baseline}
\end{subfigure} &
\begin{subfigure}[b]{0.49\textwidth}
    \includegraphics[width=\linewidth]{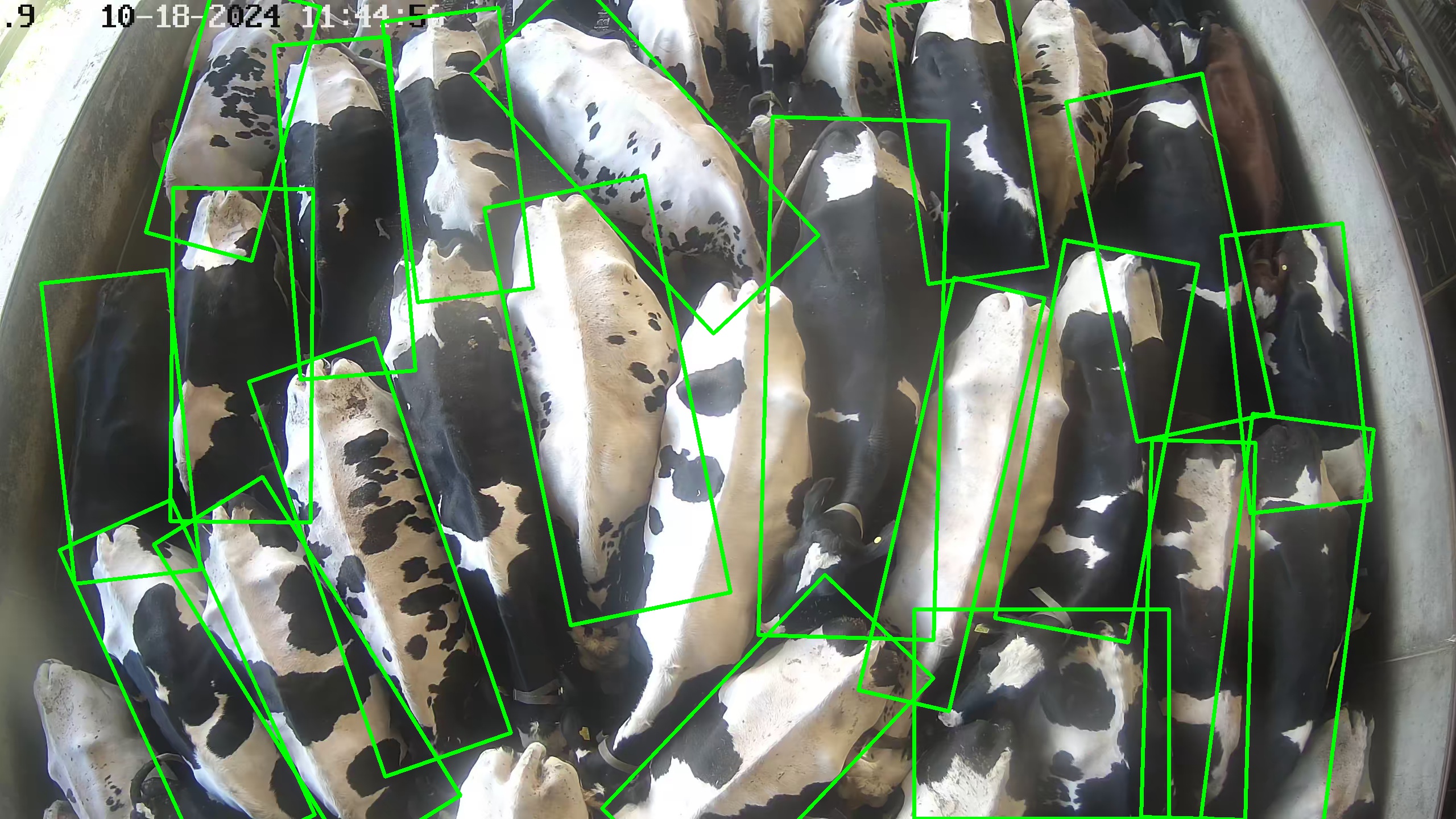}
    \caption{OWL + SAM2 \textbf{(ours)}}
\end{subfigure}
\end{tabular}
\caption{\ready{\textbf{SAM-based Species Detector Comparison.} Visualisations of detections from the standard SAM2 baseline and our OWLv2-SAM2 pipeline. Note both over-segmentations~(upper-left and middle small bounding boxes capturing only distinct coloured blobs rather than the entire cow) and missing targets~(darker cows on the right) when using the baseline.}}
\label{fig:obb-vis-sam}
\end{figure}

\ready{We have tested the accuracy of the oriented bounding boxes obtained from the standard SAM2 detector~(see Figure~\ref{fig:obb-vis-sam}), the MultiCamCows2024 species identifier~\cite{yu2025holstein} and YOLO-11 models~(see Figure~\ref{fig:obb-vis}), and that of the masks obtained from the standard SAM2 detector to analyse the performance of our species identifier component. We evaluate the mean IoU of the outputs and the matching rate~(percentage of extracted boxes that match the ground-truth above 0.7~IoU) of our framework versus the ground-truth. From the metrics evaluating semantics(Table~\ref{tab:LETOBB_table} and~\ref{tab:LET_table}), our OWLv2-SAM2 pipeline segments better than standalone SAM and surpasses the RetinaNet from MultiCamCows2024, despite not being originally designed for producing oriented bounding box segmentations. A full overview of the metrics, including target matching rates and accuracies, is available in Figures~\ref{fig:app-ex1} and~\ref{fig:app-ex2} in Appendix~A.}

\begin{table}[!htbp]
    \centering
    \begin{tabular}{l|r|r|r|r}
    \midrule
    \textbf{\textit{Backbone}} & \textbf{\textit{IoU}} & \textbf{\textit{TP}} & \textbf{\textit{Usage}} & \textbf{\textit{Matching}} \\  \textbf{\textit{}} & \textbf{\textit{}} & \textbf{\textit{Accuracy}} & \textbf{\textit{Rate}} & \textbf{\textit{Rate}} \\ \hline
    $RetinaNet$~\cite{yu2025holstein} & $0.256$ & $0.00\%$ &$1.04\%$ & $51.41\%$\\
    $SAM2$ $Baseline$~\cite{ravi2024sam2} & $0.383$ & $20.83\%$ & $26.97\%$ & $71.80\%$\\
    $OWLv2+SAM2$~$\textbf{(ours)}$ & \textbf{0.450} & \textbf{25.00\%} & \textbf{33.80\%} & \textbf{98.93\%}\\
    
    \bottomrule
    \end{tabular}
    \caption{\ready{\textbf{Oriented Bounding Box Localisation.} Shown are metrics to evaluate oriented bounding box detection results from MultiCamCows2024(RetinaNet), baseline SAM2 and our pipeline with ground truth data. \textit{IoU}: Intersection over Union with respect to ground truth versus detected bounding boxes. \textit{TP Accuracy}: Average IoU for all individuals which have been well-detected (individuals' average IoU $> 0.7$ over all frames). This metric punishes missing detections more than IoU alone.
    \textit{Usage Rate}: Fraction of total detections for which IoU $> 0.7$.
    \textit{Matching Rate}: Fraction of ground-truth bounding boxes that have a corresponding well-detected bounding box. This metric helps explain over- versus under-segmentation.}}
    \label{tab:LETOBB_table}
\end{table}

\begin{table}[!htbp]
    \centering
    \begin{tabular}{l|r|r|r|r}
    \midrule
    \textbf{\textit{Backbone}} & \textbf{\textit{IoU}} & \textbf{\textit{TP}} & \textbf{\textit{Usage}} & \textbf{\textit{Matching}} \\  \textbf{\textit{}} & \textbf{\textit{}} & \textbf{\textit{Accuracy}} & \textbf{\textit{Rate}} & \textbf{\textit{Rate}} \\ \hline
    $SAM2$ $Baseline$~\cite{ravi2024sam2} & $0.821$ & $41.67\%$ & $43.80\%$ & $46.96\%$\\
    $OWLv2+SAM2$~$\textbf{(ours)}$ & \textbf{0.898} & \textbf{70.83\%} & \textbf{75.39\%} & \textbf{94.11\%}\\
    
    \bottomrule
    \end{tabular}
    \caption{\ready{\textbf{Pixel-Level Segmentation Evaluation.} Shown are metrics to evaluate baseline SAM2 and our instance segmentations with ground truth data. Here, the same metrics are evaluated on the pixel-level masks, rather than for the oriented bounding boxes of Table~\ref{tab:LETOBB_table}.}}
    \label{tab:LET_table}
\end{table}

\subsection{Clustering Evaluation}

\ready{To further assess the quality of the automatically created dataset based on skin patterns and morphology, we have applied a UCL framework using a ResNet-50 backbone on one day of data to evaluate the clustering performance from ground-truth (manually-annotated) and automated RGB masks. For the contrastive learners trained with data from each day, we train 80$\%$ of our data and test with the remaining 20$\%$. We have applied a k-fold cross-validation with $k=5$ for the models with the same metrics to validate our results(see Table~\ref{tab:UCL_table}). From the metrics acquired, we show that our automated data-acquisition pipeline is capable of being used as the data for Re-ID tasks.}

\begin{table}[!htbp]
    \centering
    \begin{tabular}{l|r|r}
    \midrule
    \textit{\textbf{Metric}} & \textit{\textbf{Manual}} & \textit{\textbf{Automated}}\textbf{(ours)} \\ \hline
    $kNN Accuracy$ & $99.98\pm0.01\%$ & $97.77\pm0.65\%$\\
    $ARI$ & $0.974\pm0.001$ & $0.866\pm0.007$\\
    $AMI$ & $0.987\pm0.004$ & $0.944\pm0.006$\\
    $NMI$ & $0.988\pm0.004$ & $0.945\pm0.006$\\
    $HA$ & $97.73\pm1.35\%$ & $87.55\pm0.55\%$\\
    
    \bottomrule
    \end{tabular}
    \caption{\ready{\textbf{Unsupervised Contrastive Learning Clustering.} The metrics to evaluate clustering performance between manually annotated ground truth masks and the automated mask dataset enhanced with OWLv2-SAM2, with k-fold cross-validation where $k=5$. The kNN Accuracy is the accuracy of classification. The Adjusted Random Index~(ARI), Adjusted Mutual Information~(AMI), Normalised Mutual Information~(NMI) and accuracy of the Hungarian~Algorithm~(HA) represent the metrics for analysing clustering similarities based on ground truth labels and K-Means results.}}
    \label{tab:UCL_table}
\end{table}

\subsection{Re-ID Task Performance}

\begin{figure}[!htbp]
\centering
\captionsetup{justification=centering}
\begin{tabular}{p{0.32\textwidth} p{0.32\textwidth} p{0.32\textwidth}}
\begin{subfigure}[b]{0.32\textwidth}
    \includegraphics[width=\linewidth]{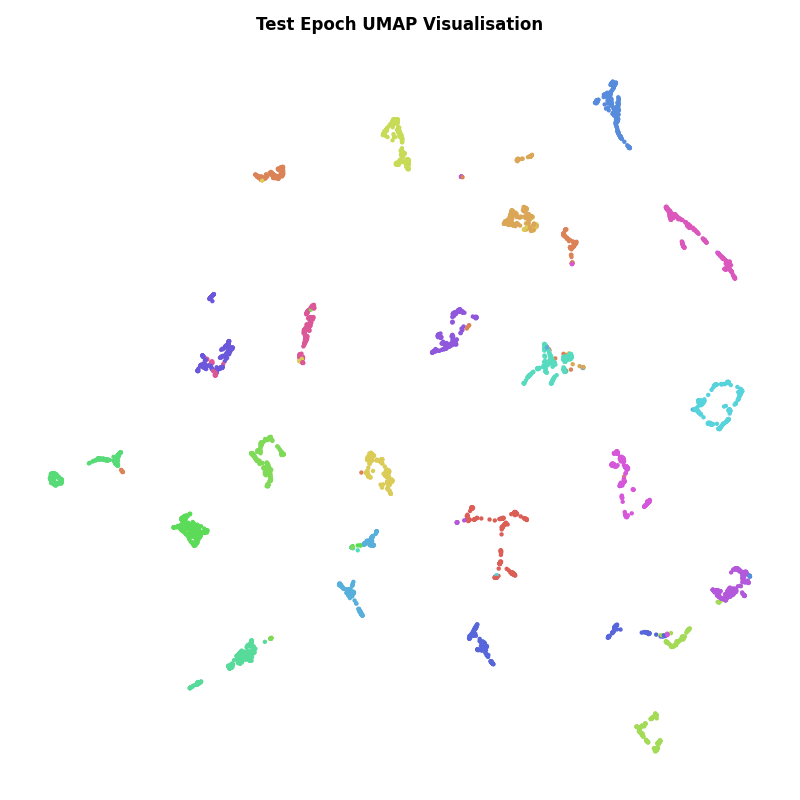}
    \caption{2024-10-18}
\end{subfigure} &
\begin{subfigure}[b]{0.32\textwidth}
    \includegraphics[width=\linewidth]{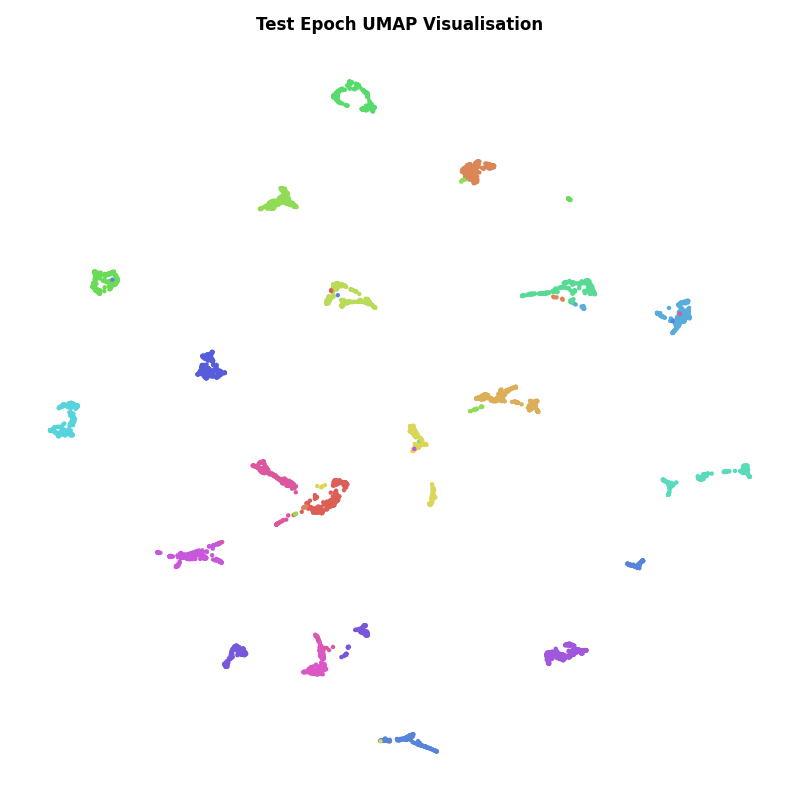}
    \caption{2024-10-31}
\end{subfigure} &
\begin{subfigure}[b]{0.32\textwidth}
    \includegraphics[width=\linewidth]{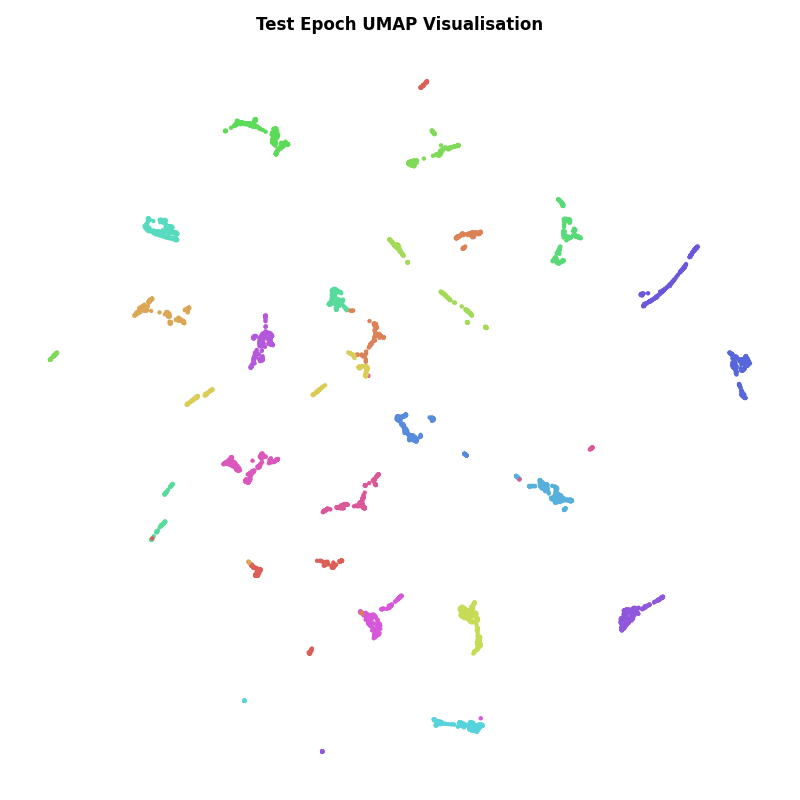}
    \caption{2024-11-01}
\end{subfigure} \\

\begin{subfigure}[b]{0.32\textwidth}
    \includegraphics[width=\linewidth]{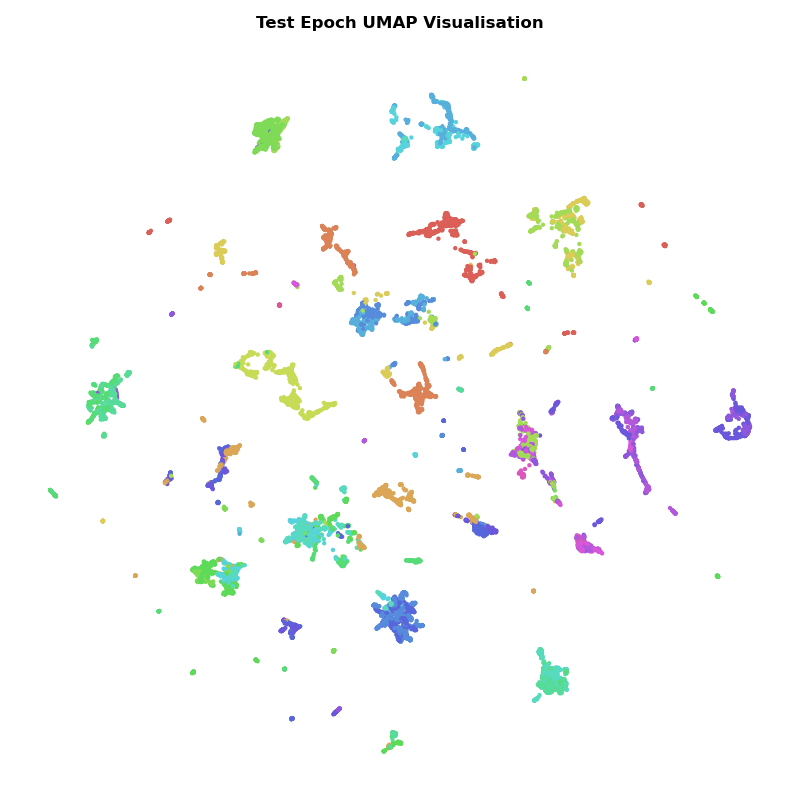}
    \caption{2024-11-02}
\end{subfigure} &
\begin{subfigure}[b]{0.32\textwidth}
    \includegraphics[width=\linewidth]{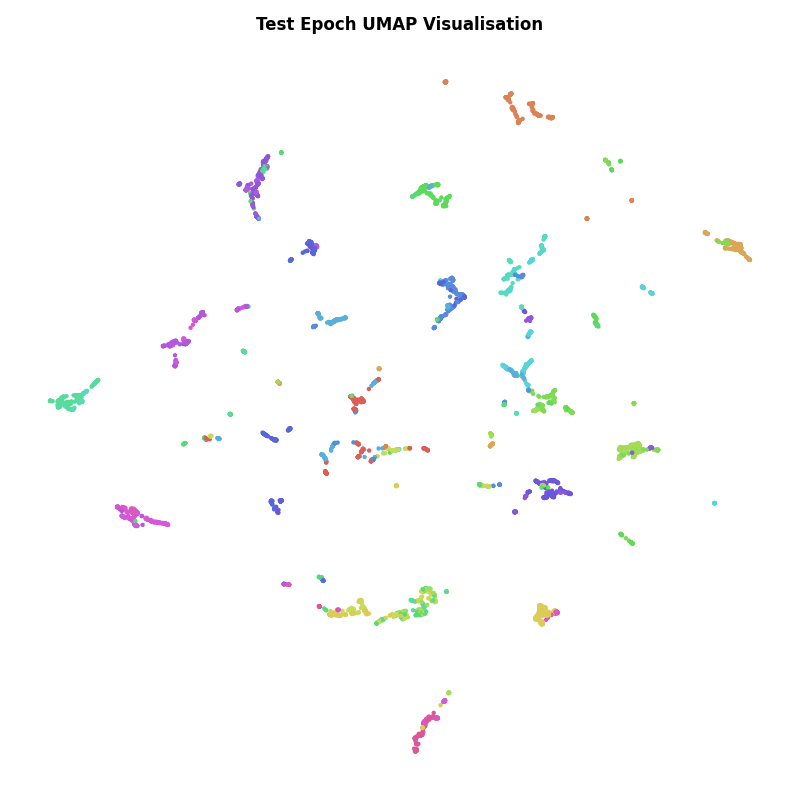}
    \caption{2024-11-03}
\end{subfigure} &
\begin{subfigure}[b]{0.32\textwidth}
    \includegraphics[width=\linewidth]{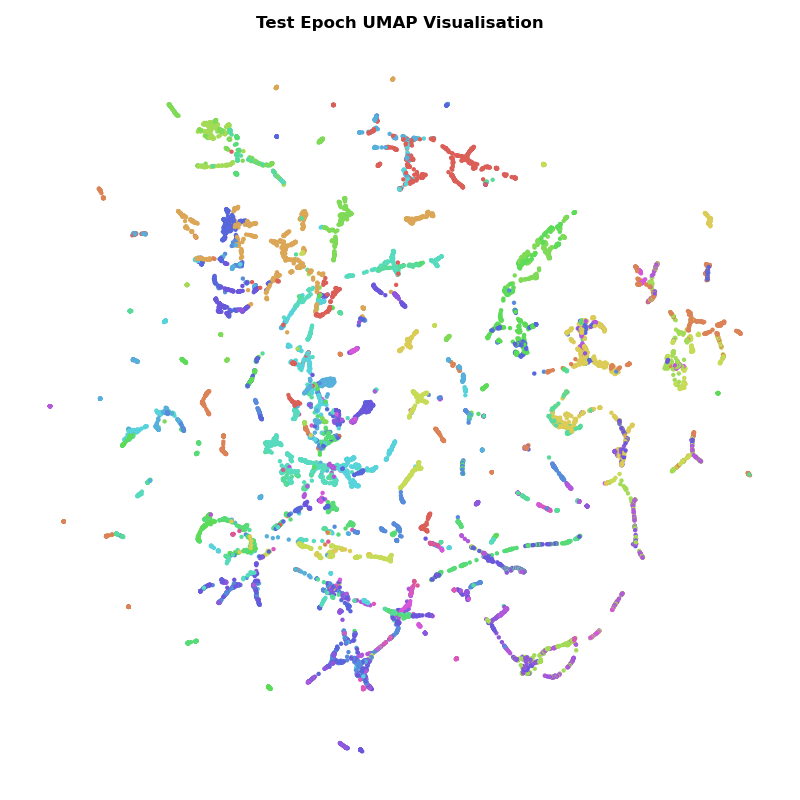}
    \caption{2024-11-04}
\end{subfigure} \\

\begin{subfigure}[b]{0.32\textwidth}
    \includegraphics[width=\linewidth]{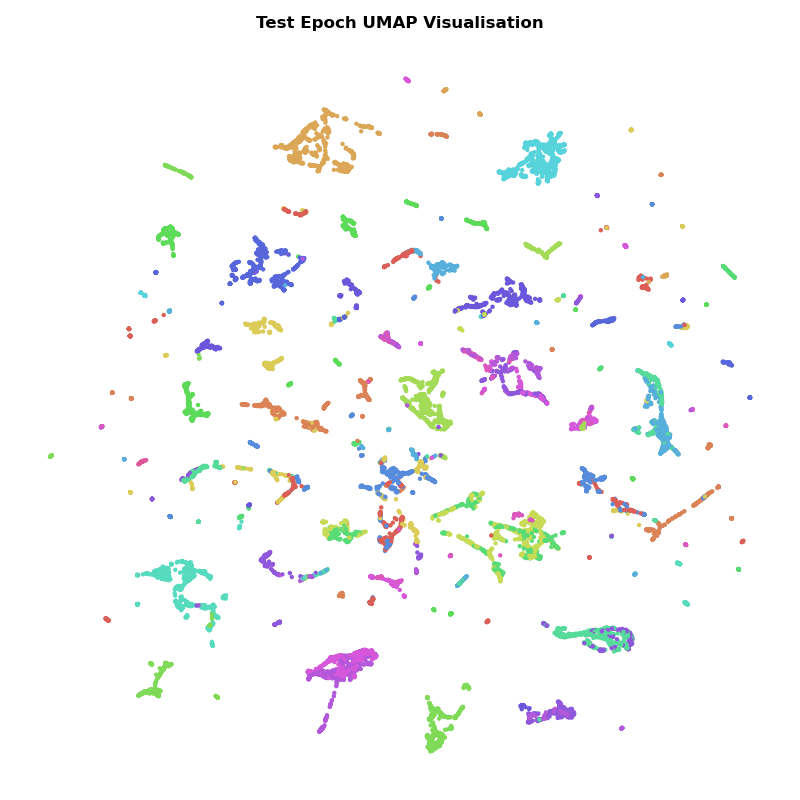}
    \caption{2024-11-05}
\end{subfigure} &
\begin{subfigure}[b]{0.32\textwidth}
    \includegraphics[width=\linewidth]{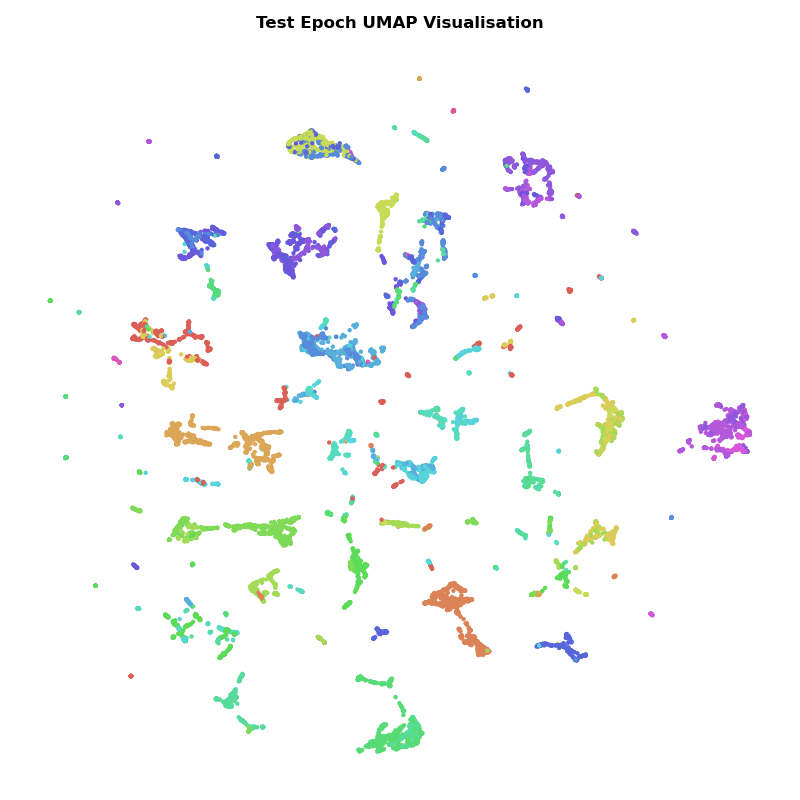}
    \caption{2024-11-06}
\end{subfigure} &
\begin{subfigure}[b]{0.32\textwidth}
    \includegraphics[width=\linewidth]{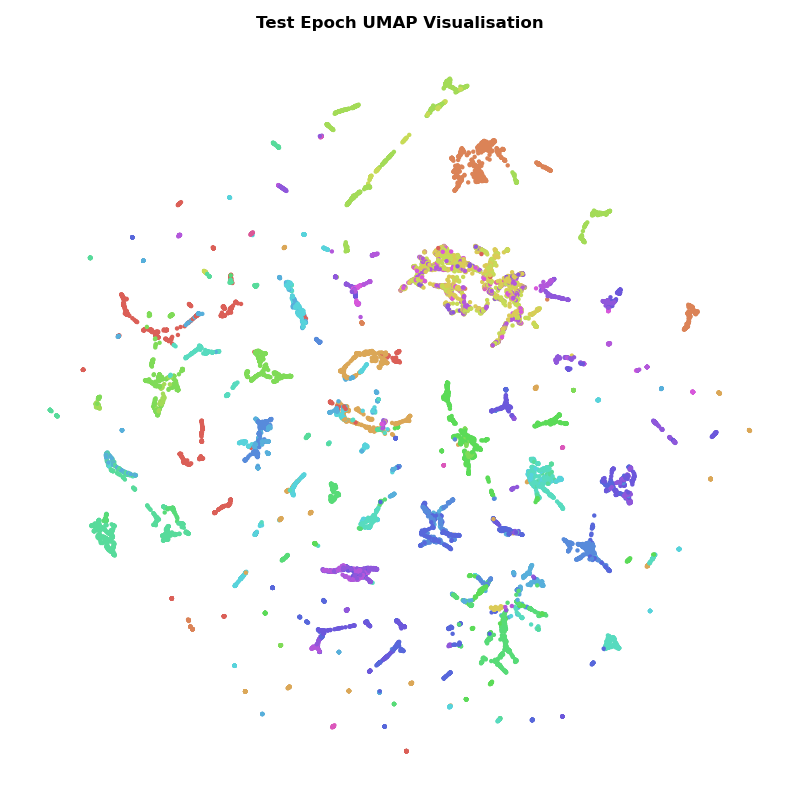}
    \caption{2024-11-07}
\end{subfigure} 

\end{tabular}
\caption{\textbf{Embeddings for Re-ID over Nine Days.} Trained on seven days of data, validated and tested on the other two days, as part of a k-fold cross-validation framework. The visualisations show testing performance, using each day's inference via the UCL-based Re-ID module. Each day shows not only different numbers of frames, but also different numbers of individuals.}

\label{fig:re-id-9-vis}
\end{figure}

\ready{To fully assess the model's ability to re-identify cows from the autonomously extracted datasets, we have tested the pipeline on an expanded dataset of data from nine days. Using k-fold cross-validation with Stochastic Gradient Descent as the optimiser for a total of 80 epochs, we retrained the UCL model on permutations of seven selected days while using the remaining two days for validation and testing~(see Table~\ref{tab:UCLINF_table} and Figure~\ref{fig:re-id-9-vis}). We have also inferred the best model obtained from the clustering evaluation phase on each of the nine days of data (Figure~\ref{fig:app-ex4}).}

\begin{table}[!htbp]
    \centering
    \begin{tabular}{l|r}
    \midrule
    \textit{\textbf{Target Dataset}} &\textit{\textbf{Test Accuracy}}\\ \hline
    $2024-10-18$ & $98.91\%$\\
    $2024-10-31$ & $99.52\%$\\
    $2024-11-01$ & $99.89\%$\\
    $2024-11-02$ & $93.96\%$\\
    $2024-11-03$ & $91.31\%$\\
    $2024-11-04$ & $88.27\%$\\
    $2024-11-05$ & $96.20\%$\\
    $2024-11-06$ & $93.84\%$\\
    $2024-11-07$ & $91.46\%$\\
    \bottomrule
    $Avg+Std$ & \textbf{$94.82\pm4.10\%$}\\
    \end{tabular}
    \caption{\ready{\textbf{kNN Accuracy on Inference and Retraining Models of Unseen Data.} Using a customised k-fold cross-validation with $k=9$, the \textit{Test Accuracy} uses the corresponding \textit{target dataset} for inference. For each instance, its previous day was used as the validation set, and the remaining seven days were used for training. (e.g. The first row metric is obtained from a model instance trained with data from 2024-10-31 to 2024-11-06, validated on data from 2024-11-07, and tested on the data from 2024-10-18.)}}
    \label{tab:UCLINF_table}
\end{table}

\newpage

\section{Discussion}
\label{sec:discussions}

\subsection{Bounding Box Detection Accuracy}

\ready{From both localisation and clustering evaluation, the refined OWLv2-SAM2 model consistently outperforms the baseline SAM2 model when used without manual localisation. While the stand-alone SAM2 model can segment based on foreground-background colour differences, it lacks semantic understanding of part-whole relationships if not trained with specific data, such as whether disjoint feature regions belong to the same cow. Consequently, SAM often yields either undersegmentation---grouping multiple cows into a single mask---or oversegmentation---splitting one cow into fragmented patterns. Both issues hinder the downstream analysis of individual health, behaviour and the inter-animal interactions of densely packed cattle.}

\ready{In addition, as shown in Tables~\ref{tab:LETOBB_table} and \ref{tab:LET_table}, although OWL-enhanced SAM achieves higher average IoU, accuracy and match rates, it still contains residual over- and under-segmentations during autonomous inference. Nevertheless, its outputs align more closely with our manually labelled ground truths compared to the baseline SAM2 model, indicating improved robustness in instance-aware segmentation.}

\ready{As shown in Figure~\ref{fig:comp-perf1} and Figure~\ref{fig:comp-perf2}, grounded models such as GroundingDINO and GroundedSAM struggle to segment individual cows without retraining. GroundingDINO, when used without image-specific fine-tuning, tends to merge all instances into a single undifferentiated region. Similarly, GroundedSAM, when provided with a generic prompt such as `cow', fails to distinguish between individuals, producing an oversegmented outline of grouped cows. In contrast, with the same single-word prompt `cow', the pretrained OWLv2 model demonstrates a greater capacity to distinguish individual cows within a dense group. Additionally, we also used other prompts for GroundedSAM and OWLv2, such as `all instances of cows among a group' or `30 individual cows in the densely packed group', to extract features. But with these longer prompts of general statements, the models fail to detect or segment completely. We hypothesise that these phenomena stem from the manner in which OWLv2 encodes the semantic concept of `cow' during pre-training, representing it at the instance level rather than as a collective class. As a result, even without additional fine-tuning, it provides coarse yet instance-aware segmentation outputs. This behaviour reflects a limitation in open-vocabulary models like GroundingDINO and GroundedSAM, where a single-word prompt such as `cow' is insufficient to guide precise instance-level segmentation of each target. Their large-scale vocabulary training can dilute the specificity of such terms, requiring more detailed input prompts or additional supervision. In practical terms, this increases the need for human-in-the-loop annotation when applying these models to species identification tasks. The necessity of composing multi-word queries to localise each individual not only increases input complexity but also demands a true understanding of the number of cows present.}

\ready{The assistance of OWLv2 brings faster and more accurate segmentations for the subsequent SAM2 inference, enabling more accurate body pattern recognition using UCL. Utilising the bounding box data restricts the regions of interest per target, enabling SAM2 to infer within a smaller, more accurate region, thereby reducing computation whilst increasing overall accuracy. SAM2 can now better track targets in videos due to the OWLv2 localisations.}

\subsection{Automated dataset for Re-ID tasks}

\ready{We have used UCL to measure the quality of the clustering on the automated dataset with human-annotated ground truths(Table~\ref{tab:UCL_table}) for the dataset's Re-ID capability. We see that kNN works well for classification compared to the ground-truth ($97.77\%$ vs. $99.98\%$). Our metrics for clustering performance also show strong consistency between embeddings and predicted labels(ARI: $0.866$ vs. $0.974$; HA: $87.55\%$ vs. $97.73\%$). We argue that the metrics above, alongside the visualisations in Appendix~B confirms that the automated dataset contains well-separated and semantically meaningful object masks.}


\ready{For Re-ID, trained with multiple days and inferred on unseen data, the k-fold instances demonstrate promising kNN accuracies($94.82\%$). These results mirror the observed performance of the unsupervised contrastive learner via UMAP visualisations(Figure~\ref{fig:re-id-9-vis}). The results demonstrate the capability of our UCL pipeline for re-identifying individuals in grouped cows using unseen data. The Re-ID model's dataset is acquired from our automated mask extraction pipeline, removing humans completely from the loop. We suggest that the failure cases of our Re-ID originate from cases of bad segmentations. Therefore, improvements in data acquisition and preprocessing, such as filtering out undersegmented or oversegmented masks, could further enhance the model's sensitivity to local distinctions and lead to better cluster fidelity.}

\ready{Accuracy ranges from $88.27\%$ to $99.52\%$ due to different movement activities across the days of the experiment: rapid movements correlate with lower accuracy due to shorter tracks in front of the cameras with less information per cow. Despite this, our pipeline achieves relatively high-levels of re-identification without external labelling. This includes both the automated data acquisition phase with OWLV2-SAM2 and the contrastive learning phase. The zero labelling requirement is, in part, due to the knowledge that each individual appears only once per image.}

\subsection{Future Work}



\textbf{Depth Feature as an Alternative.} \ready{In addition to using visual features extracted from open-vocabulary frameworks for text-vision matching, there are other, purely visual-based methodologies. The Depth Anything model~\cite{yang2024depth}, trained with no information other than the encoded visual representation, acquires both colour and depth information of targets with respect to the cameras. Augmenting colour with depth allows for easier segmentation and tracking, as distance information is encoded. Even though a customised Depth Anything encoder may need to be pre-trained to cover all permutations of inputs, it allows completely autonomous tracking of camouflaged targets. It is also free from text prompts that may contain potential inexactness in the target description.}

\textbf{Inter-module Enhancements.} \ready{The framework, with both of its main components fully operational on their own, also allows further refinement of backbone weights and augmentations. Between the transformation from bounding box outputs to SAM2 inputs, intermediate matching of outputs can be performed to track camouflaged targets in videos.
GroundedSAM, though unable to extract individual cows in our scenario, might be further restructured into a better replacement of the standard SAM2 model. This would, however, require correlating the text prompts and OWL inferences.
In this manner, text prompts can be used for both OWLv2 and the SAM2 component, enabling better tracking. The latest SAM3~\cite{carion2025sam} also promises to work as an upgrade to SAM2, offering superior segmentation performance and conceptual understanding of text prompts over SAM2.}

\textbf{Towards Camouflaged Object Detection.} \ready{Related research in camouflaged object detection(COD) addresses visual ambiguity similar to our challenge in detecting dazzle patterns. This may inform future adaptations of our pipeline. Imagery and video COD toolkits specialise in enhanced learning of feature representations from metadata, such as the sizes and aspect ratios of targets~\cite{galun2003texture}, ad-hoc colour contrasts~\cite{siricharoen2010robust, kavitha2011efficient} and 3D convexity~\cite{tankus1998detection, tankus2001convexity, pan2011study}, or features preprocessed by deep learning~\cite{sun2021context, chen2022camouflaged, yin2024camoformer, pang2024open}. Well-known for their biologically unique body patterns of black and white blobs, detecting and segmenting individual Holstein-Friesian cows within groups face dazzle pattern issues similar to general COD. Therefore, the occurrences of over- and undersegmentation from our pipeline may be alleviated with the addition of such metadata.}

\section{Conclusion}
\label{sec:conclusions}

\ready{We introduce a simple yet effective framework that leverages OWLv2 and SAM2 to detect each Holstein-Friesian cow when they stay close together in crowds. Without any retraining and with the input of simple text, the framework produces refined mask segmentations with a~$48\%$ and~$27\%$ boost in localisation and feature extraction performance, compared with RetinaNet and standard SAM2 baseline detectors. Removing the need for manual correction and explicit identity supervision, the pipeline is capable of reaching a $95\%$ accuracy regarding Re-ID in a real-world dataset covering nine days of data. Furthermore, the pipeline itself is training-free~(and hence easily transferable to new cameras or farms) and also adaptable to other downstream tasks. We conclude that automated visual Re-ID in densely crowded settings is not only feasible, but also practically implementable without human labelling~--~even in gathering areas (e.g. before milking) of working farms where Re-ID of animals can now be performed.}

\newpage
\appendix

\section{Additional Localisation Visualisations}

\begin{figure}[!htbp]
\centering
\captionsetup{justification=centering}
\begin{tabular}{p{0.32\textwidth} p{0.32\textwidth} p{0.32\textwidth}}
\begin{subfigure}[b]{0.32\textwidth}
    \includegraphics[width=\linewidth]{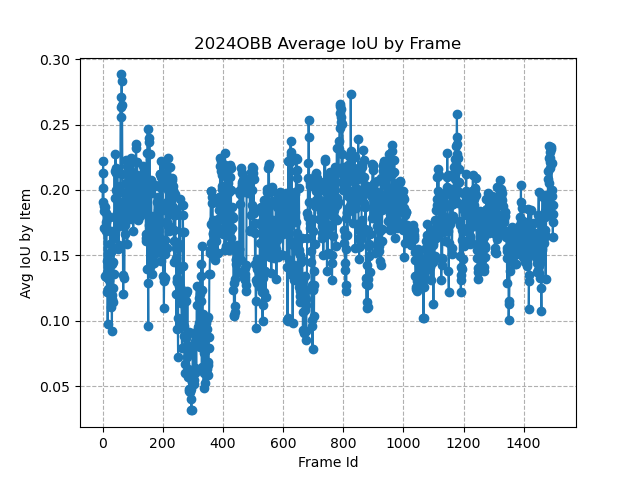}
    \caption{RetinaNet~\cite{yu2025holstein}}
\end{subfigure} &
\begin{subfigure}[b]{0.32\textwidth}
    \includegraphics[width=\linewidth]{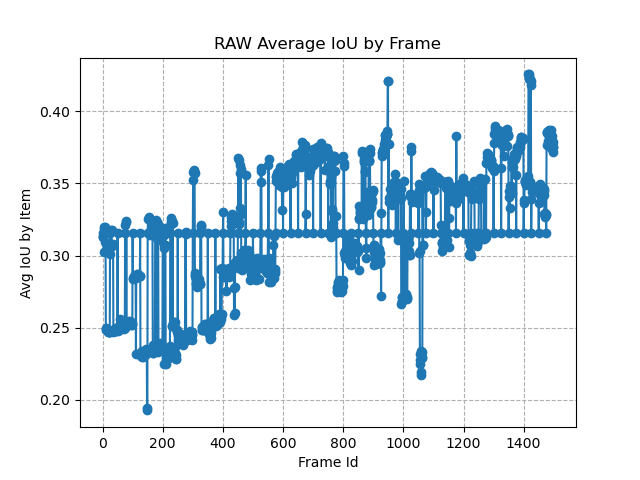}
    \caption{Baseline SAM2~\cite{ravi2024sam2}}
\end{subfigure} &
\begin{subfigure}[b]{0.32\textwidth}
    \includegraphics[width=\linewidth]{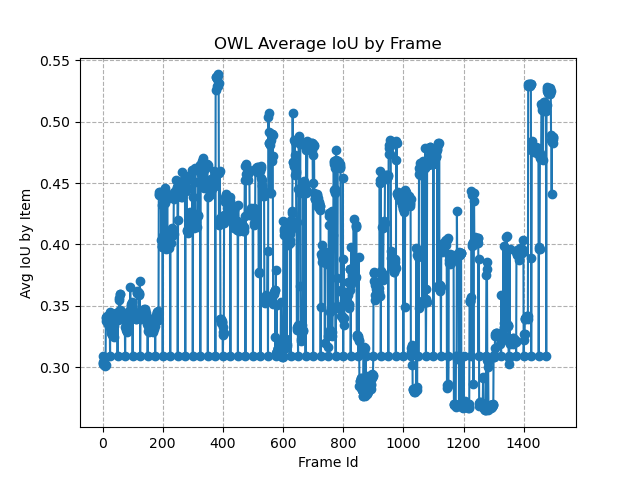}
    \caption{OWL + SAM2~\textbf{(ours)}}
\end{subfigure} \\

\begin{subfigure}[b]{0.32\textwidth}
    \includegraphics[width=\linewidth]{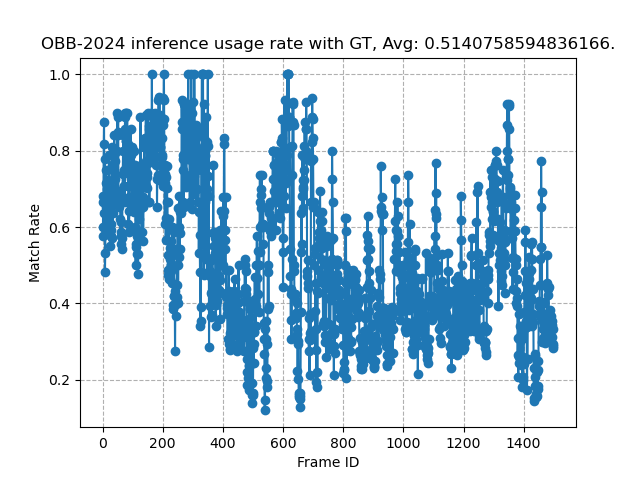}
    \caption{RetinaNet~\cite{yu2025holstein}}
\end{subfigure} &
\begin{subfigure}[b]{0.32\textwidth}
    \includegraphics[width=\linewidth]{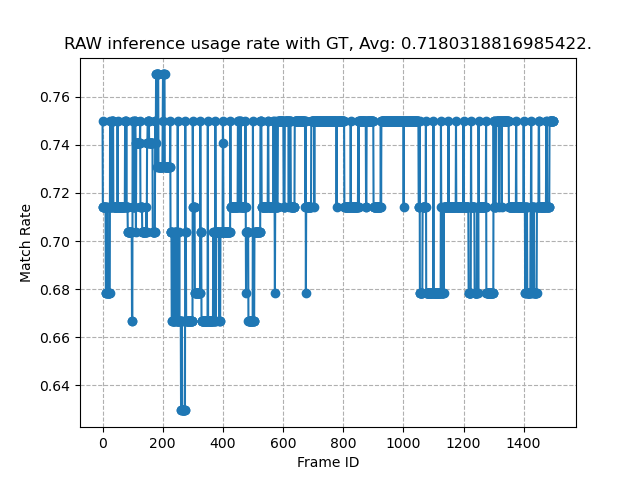}
    \caption{Baseline SAM2~\cite{ravi2024sam2}}
\end{subfigure} &
\begin{subfigure}[b]{0.32\textwidth}
    \includegraphics[width=\linewidth]{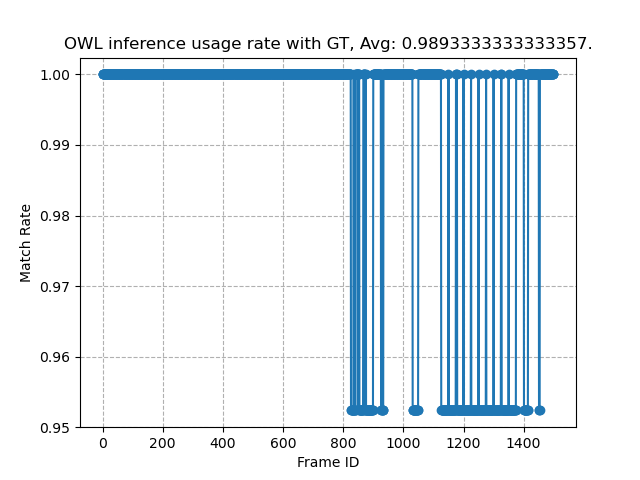}
    \caption{OWLv2 + SAM2~\textbf{(ours)}}
\end{subfigure}

\end{tabular}
\caption{\ready{\textbf{Oriented Bounding Box Results Metrics.} Additional metrics for oriented bounding boxes using a part of our new dataset. We show results for RetinaNet-based, baseline SAM2 and our OWL2-enhanced SAM2 detection. From a video with a total of $1500$ frames, the~\textit{(upper)} row represents the average IoU of the corresponding segmentation per video frame. These are based on the manually labelled ground truths of a total of $24$ individuals. The~\textit{(lower)} row demonstrates the ratio of well-detected bounding boxes, a frame-level Usage Rate.}}
\label{fig:app-ex1}
\end{figure}

\begin{figure}[!htbp]
\centering
\captionsetup{justification=centering}
\begin{tabular}{p{0.49\textwidth} p{0.49\textwidth}}
\begin{subfigure}[b]{0.49\textwidth}
    \includegraphics[width=\linewidth]{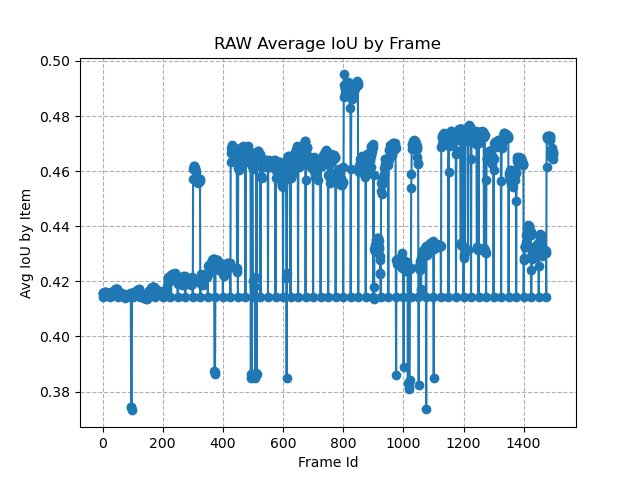}
    \caption{Baseline SAM2~\cite{ravi2024sam2}}
\end{subfigure} &
\begin{subfigure}[b]{0.49\textwidth}
    \includegraphics[width=\linewidth]{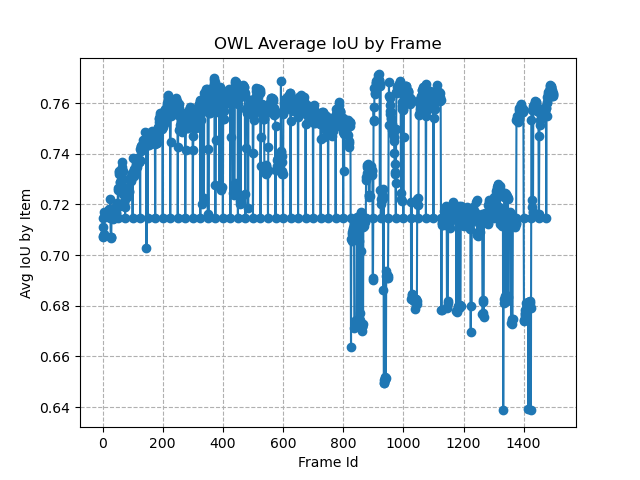}
    \caption{OWL + SAM2~\textbf{(ours)}}
\end{subfigure} \\

\begin{subfigure}[b]{0.49\textwidth}
    \includegraphics[width=\linewidth]{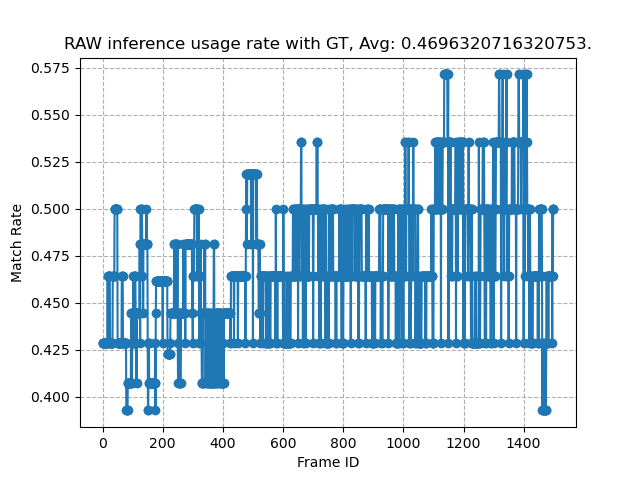}
    \caption{Baseline SAM2~\cite{ravi2024sam2}}
\end{subfigure} &
\begin{subfigure}[b]{0.49\textwidth}
    \includegraphics[width=\linewidth]{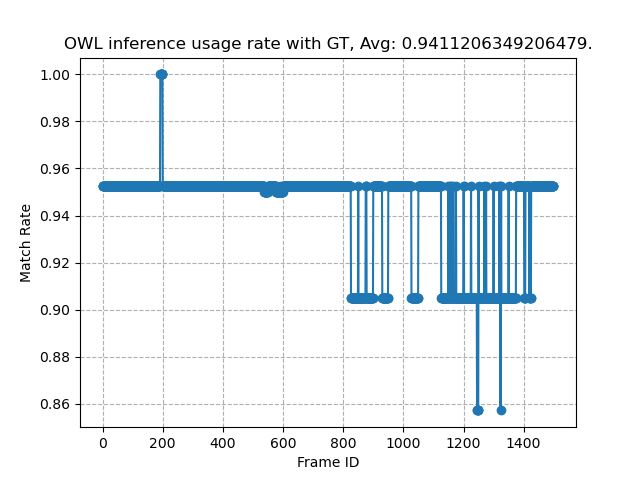}
    \caption{OWL + SAM2~\textbf{(ours)}}
\end{subfigure}

\end{tabular}
\caption{\ready{\textbf{Mask Segmentations Metrics.} Additional metrics for masks using a part of our new dataset. We show results for baseline SAM2 and our OWL2-enhanced SAM2 detection. From a video with a total of $1500$ frames, the~\textit{(upper)} row represents the average IoU of the corresponding segmentation per video frame. These are based on the manually labelled ground truths of a total of $24$ individuals. The~\textit{(lower)} row demonstrates the ratio of well-detected bounding boxes(Usage Rate) per frame.}}
\label{fig:app-ex2}
\end{figure}

\newpage
\section{Unsupervised Contrastive Learning Embeddings}

\begin{figure}[!htbp]
\centering
\captionsetup{justification=centering}
\begin{tabular}{p{0.32\textwidth} p{0.32\textwidth} p{0.32\textwidth}}
\begin{subfigure}[b]{0.32\textwidth}
    \includegraphics[width=\linewidth]{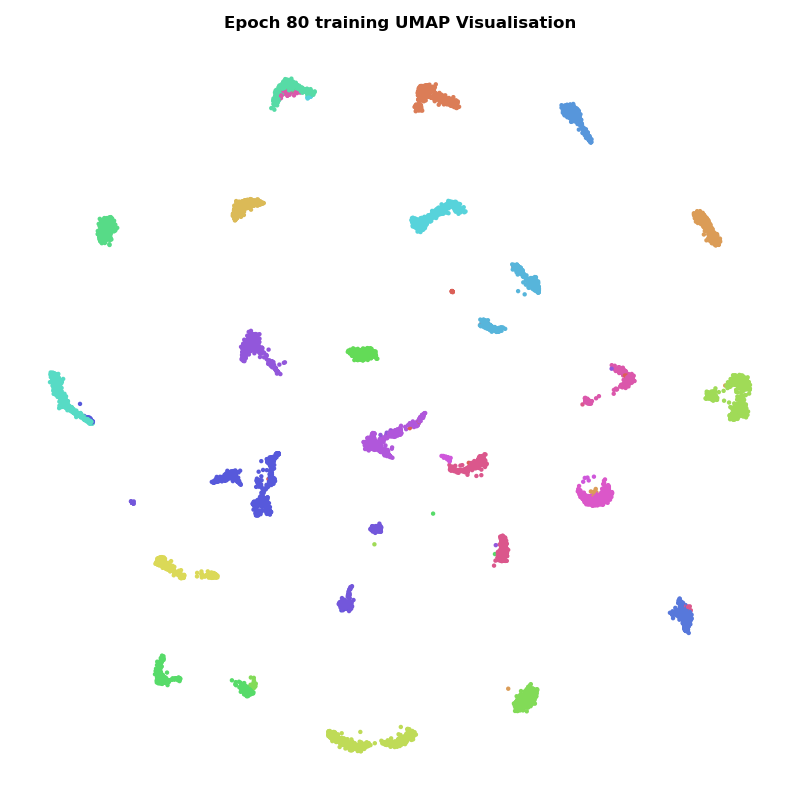}
    \caption{Manual set trained embedding with K-Means labels}
\end{subfigure} &
\begin{subfigure}[b]{0.32\textwidth}
    \includegraphics[width=\linewidth]{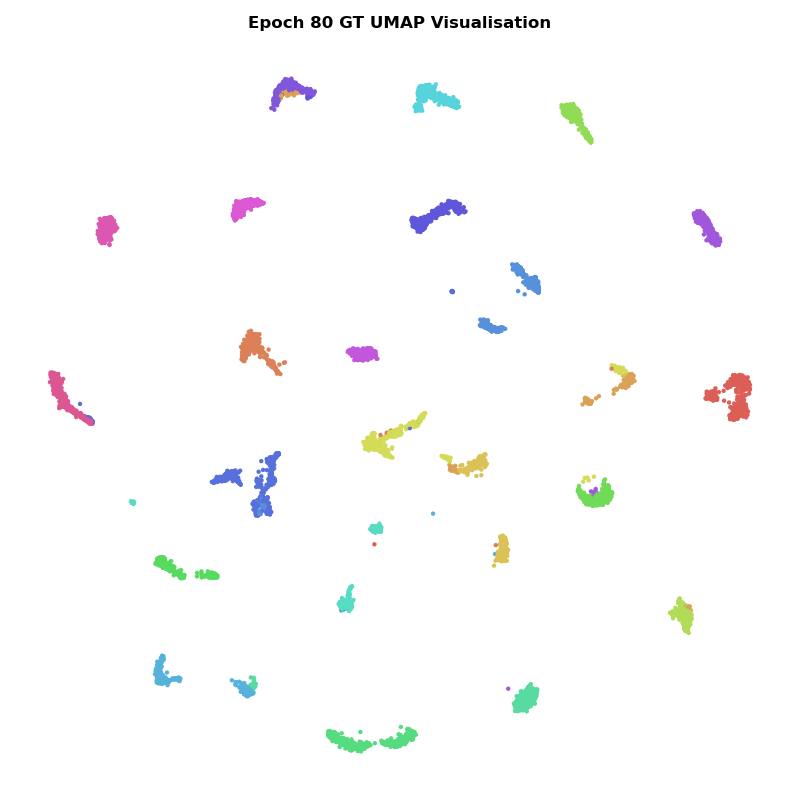}
    \caption{Manual set trained embedding with ground truth labels}
\end{subfigure} &
\begin{subfigure}[b]{0.32\textwidth}
    \includegraphics[width=\linewidth]{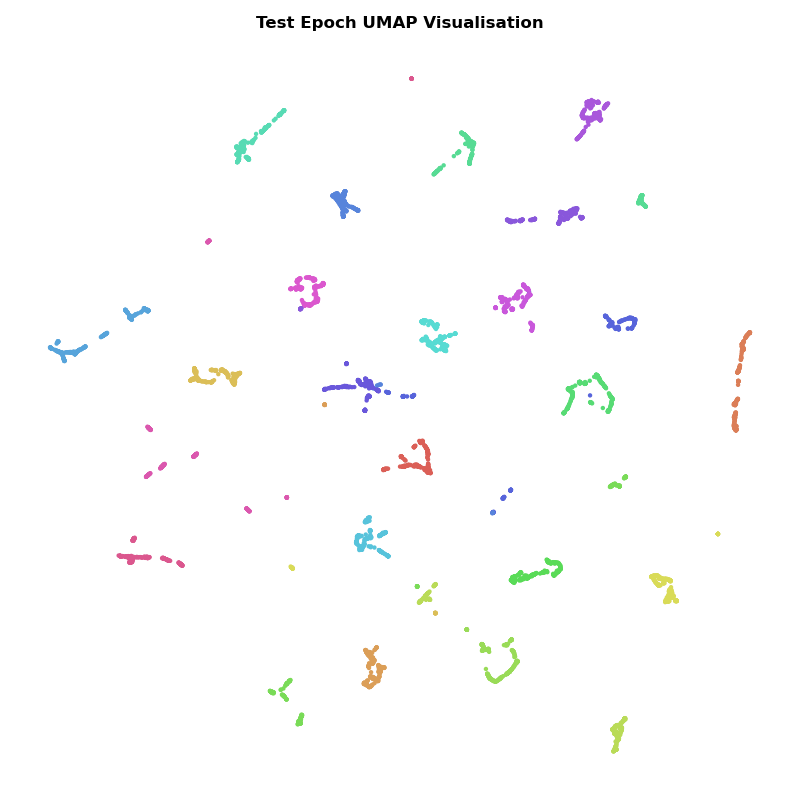}
    \caption{Best embedding of Manual set in test mode}
\end{subfigure} \\

\begin{subfigure}[b]{0.32\textwidth}
    \includegraphics[width=\linewidth]{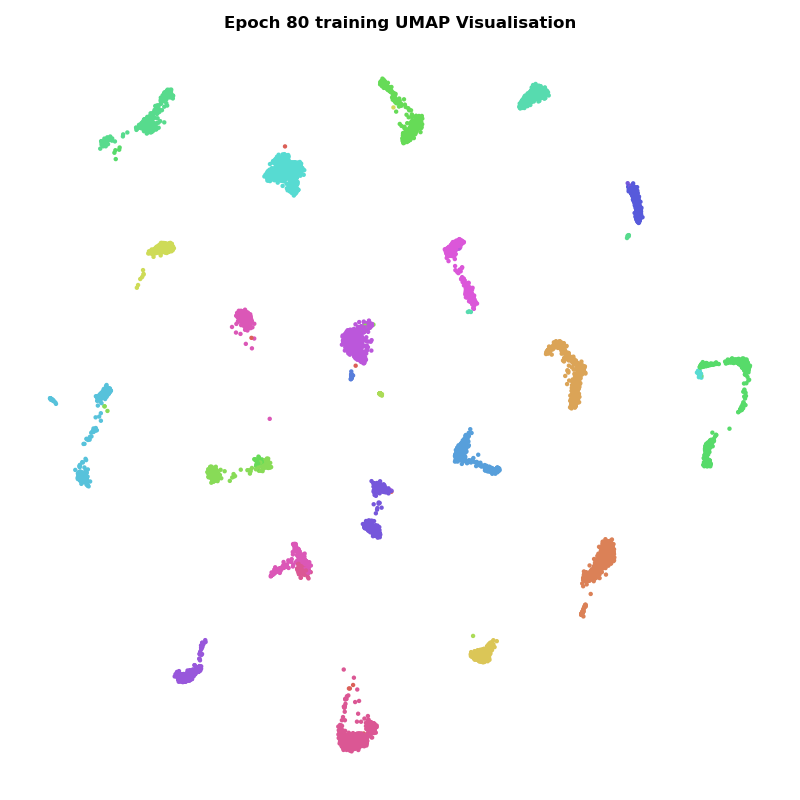}
    \caption{OWLv2-enhanced set trained embedding with K-Means labels}
\end{subfigure} &
\begin{subfigure}[b]{0.32\textwidth}
    \includegraphics[width=\linewidth]{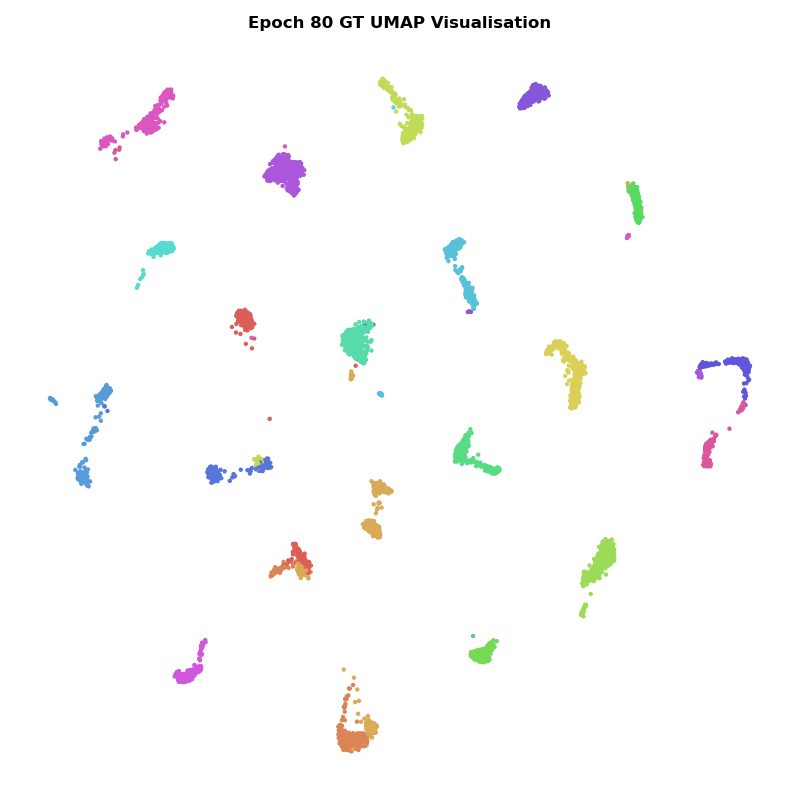}
    \caption{OWLv2-enhanced set trained embedding with ground truth labels}
\end{subfigure} &
\begin{subfigure}[b]{0.32\textwidth}
    \includegraphics[width=\linewidth]{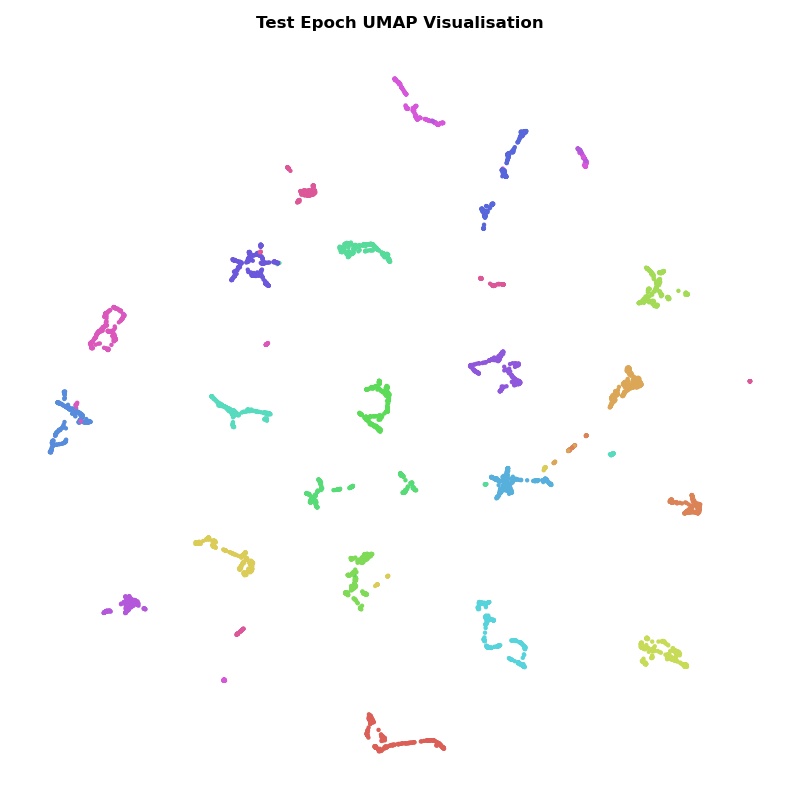}
    \caption{Best embedding of OWLv2-enhanced set in test mode}
\end{subfigure} 

\end{tabular}
\caption{\ready{\textbf{Embedding Visualisations.} Embeddings of our hand-labelled (single day worth of) ground truth versus automatically extracted masks. The~\textit{(upper)} row represents the results from the hand-labelled dataset, while the~\textit{(lower)} from our automated framework. (Left) shows the embedding, for the training set, coloured with the best K-Means proposed labels. \textit{(Middle)} column is the same embedding but coloured with the ground truth labels.\textit{(Right)} shows the test set coloured using the ground truths. Ideally, they would all be the same, with every cluster having a single colour.}}

\label{fig:app-ex3}
\end{figure}
\newpage

\begin{figure}[!htbp]
\centering
\captionsetup{justification=centering}
\begin{tabular}{p{0.32\textwidth} p{0.32\textwidth} p{0.32\textwidth}}
\begin{subfigure}[b]{0.32\textwidth}
    \includegraphics[width=\linewidth]{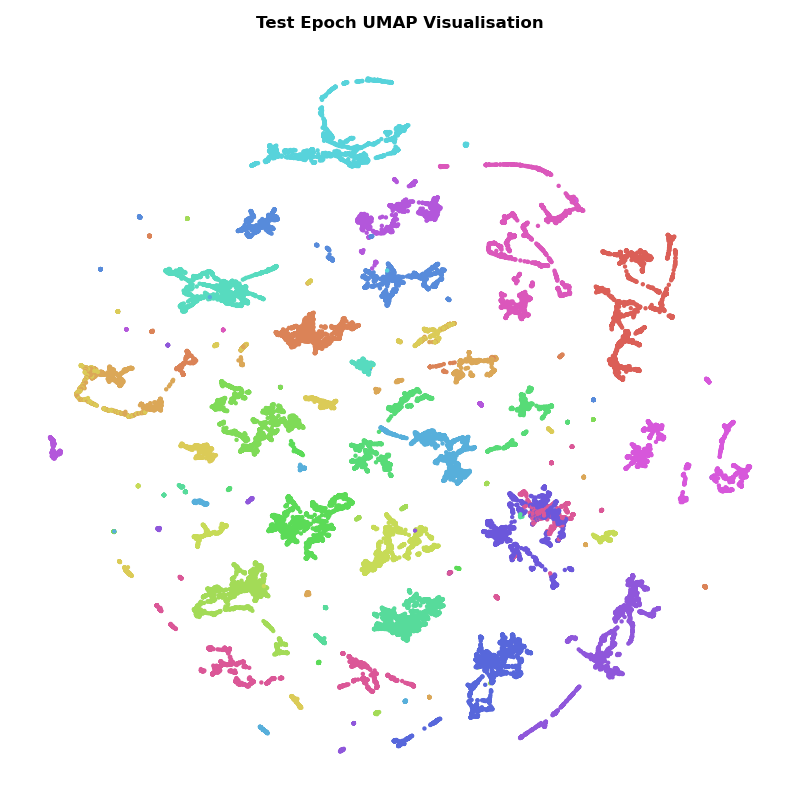}
    \caption{2024-10-18}
\end{subfigure} &
\begin{subfigure}[b]{0.32\textwidth}
    \includegraphics[width=\linewidth]{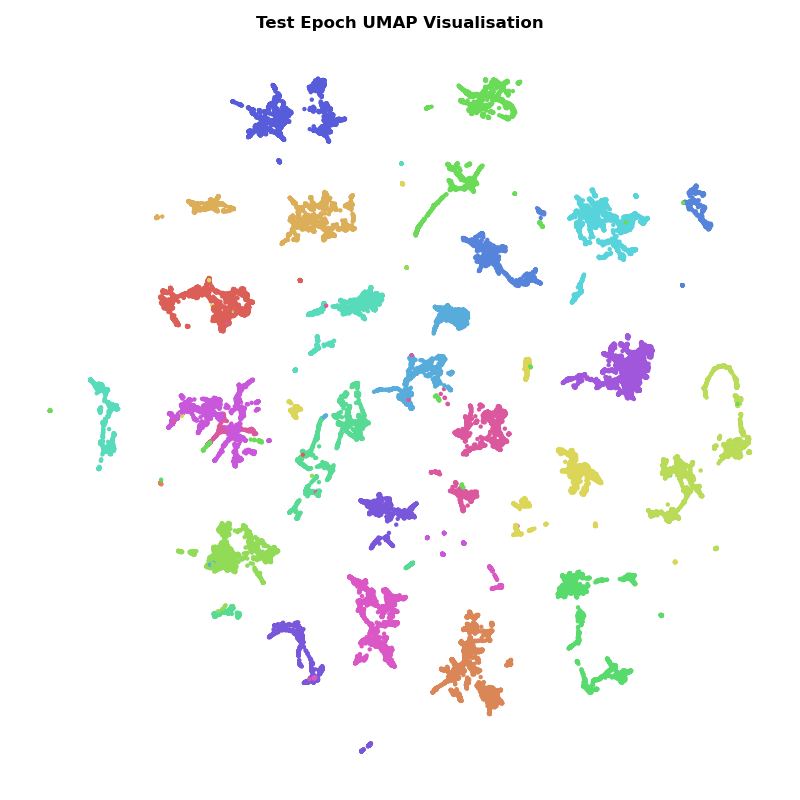}
    \caption{2024-10-31}
\end{subfigure} &
\begin{subfigure}[b]{0.32\textwidth}
    \includegraphics[width=\linewidth]{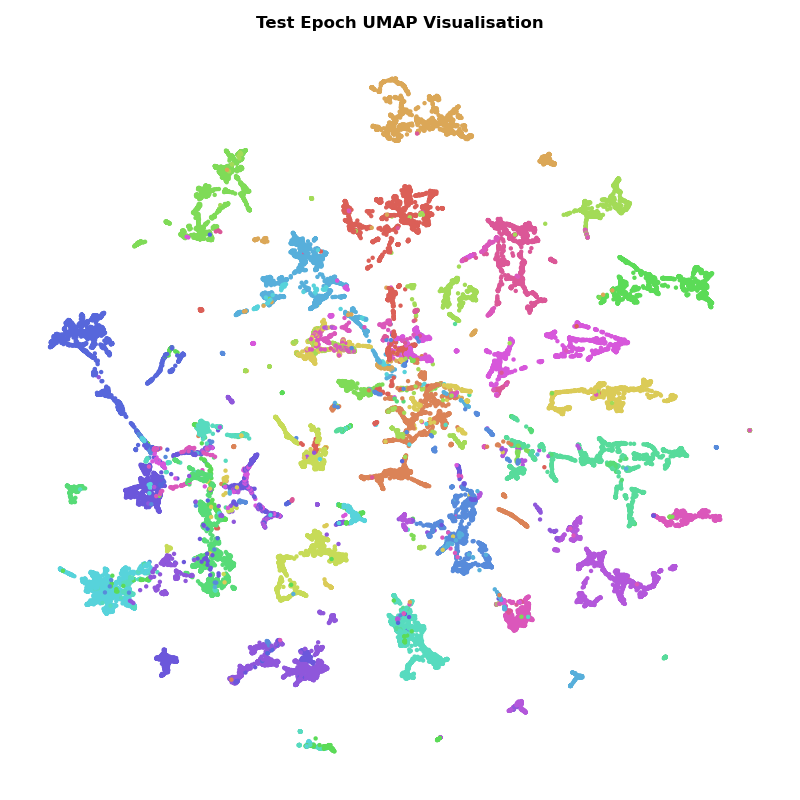}
    \caption{2024-11-01}
\end{subfigure} \\

\begin{subfigure}[b]{0.32\textwidth}
    \includegraphics[width=\linewidth]{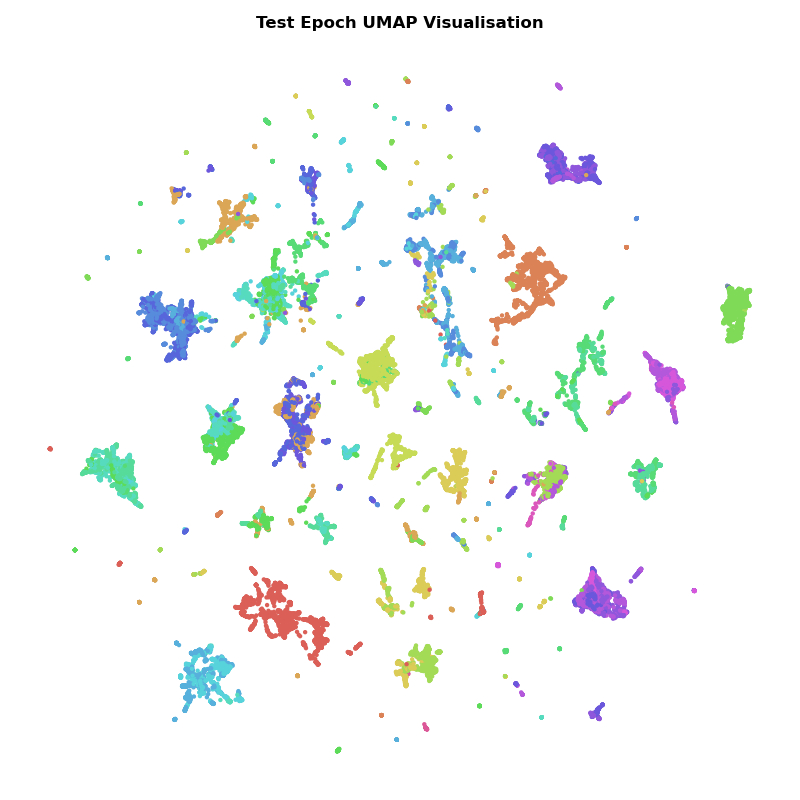}
    \caption{2024-11-02}
\end{subfigure} &
\begin{subfigure}[b]{0.32\textwidth}
    \includegraphics[width=\linewidth]{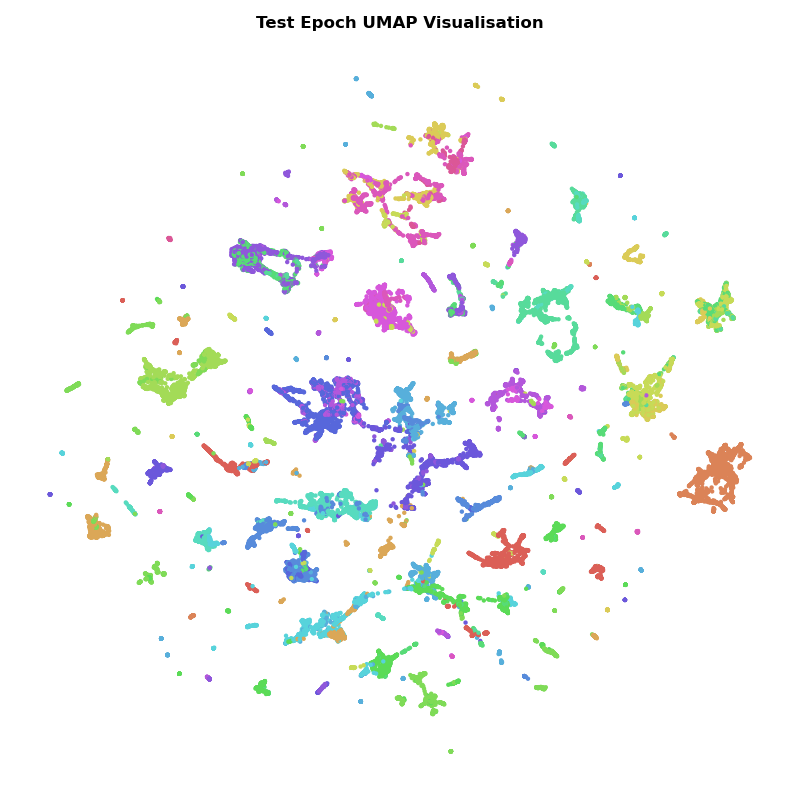}
    \caption{2024-11-03}
\end{subfigure} &
\begin{subfigure}[b]{0.32\textwidth}
    \includegraphics[width=\linewidth]{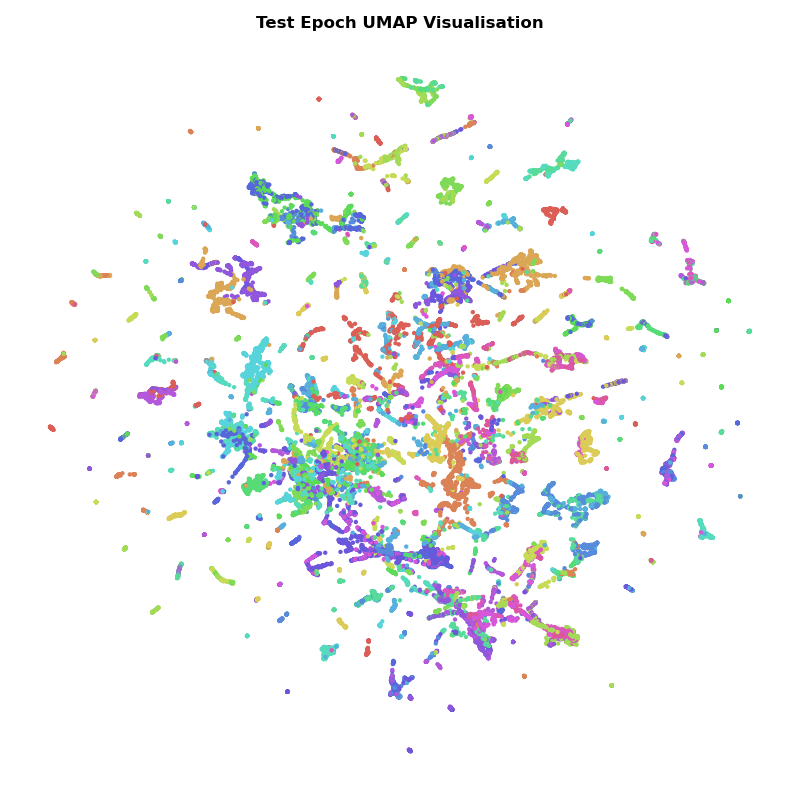}
    \caption{2024-11-04}
\end{subfigure} \\

\begin{subfigure}[b]{0.32\textwidth}
    \includegraphics[width=\linewidth]{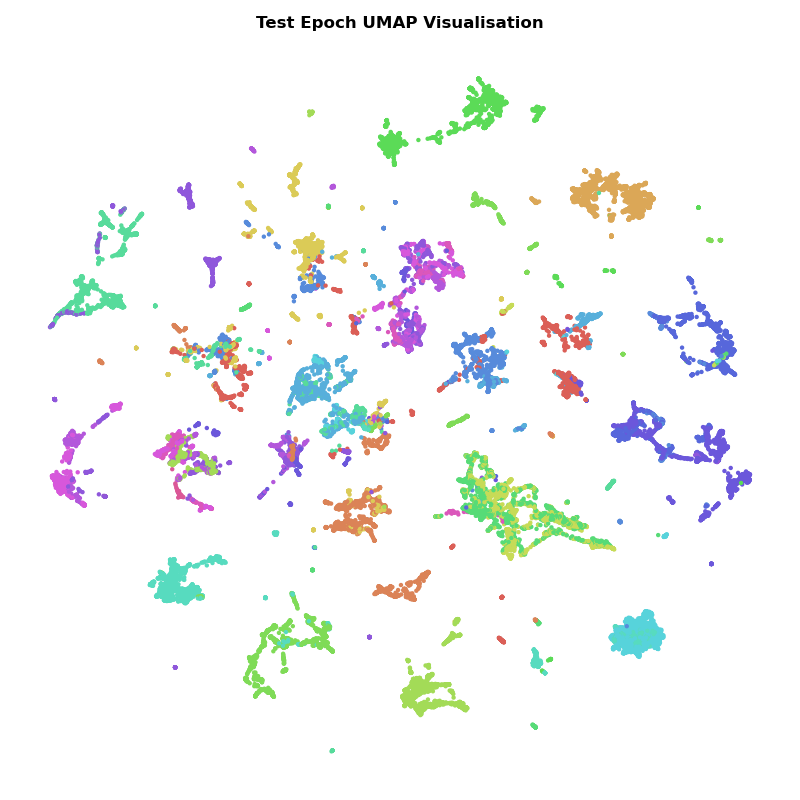}
    \caption{2024-11-05}
\end{subfigure} &
\begin{subfigure}[b]{0.32\textwidth}
    \includegraphics[width=\linewidth]{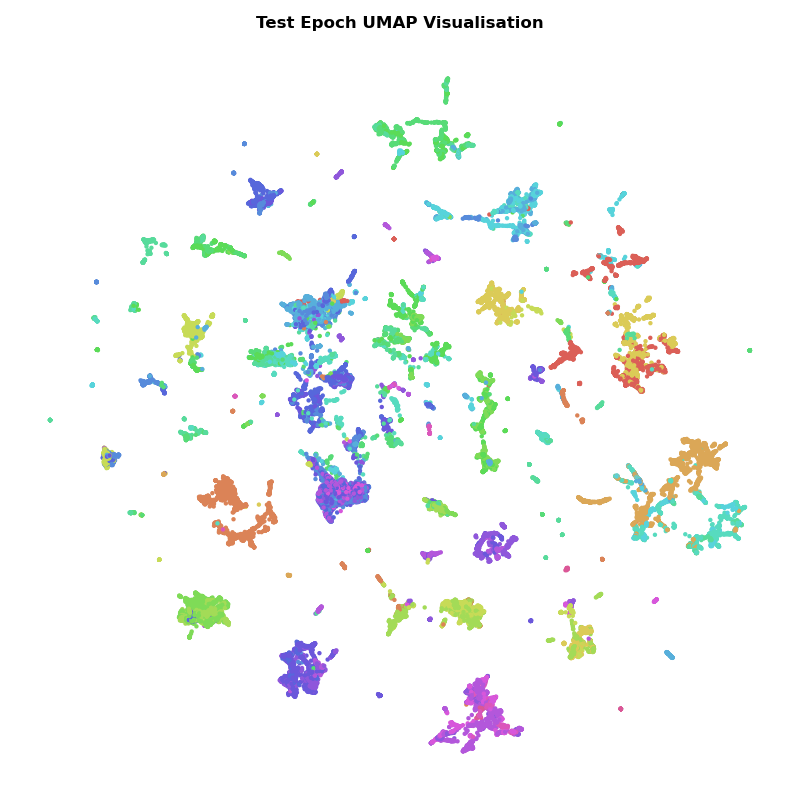}
    \caption{2024-11-06}
\end{subfigure} &
\begin{subfigure}[b]{0.32\textwidth}
    \includegraphics[width=\linewidth]{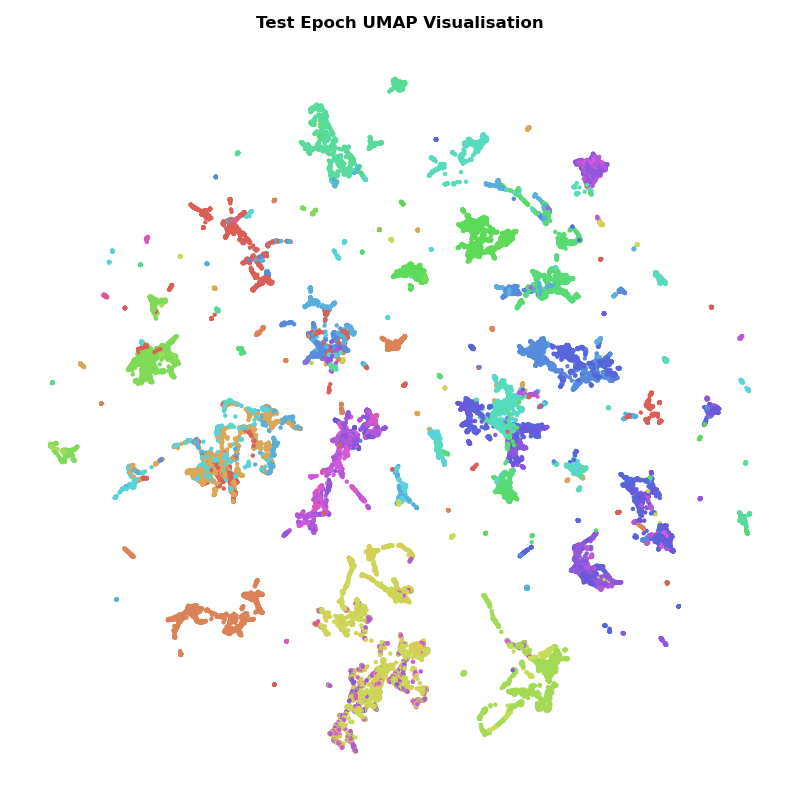}
    \caption{2024-11-07}
\end{subfigure} 

\end{tabular}
\caption{\ready{\textbf{Visualisations of Single-Day UCL Embeddings.} Embeddings showing clustering for each day in turn. The dataset is obtained with our automated OWLv2-SAM2 pipeline. The Figure demonstrates that Re-ID may be possible using only a single day's data.}}

\label{fig:app-ex4}
\end{figure}

\begin{table}[!htbp]
    \centering
    \begin{tabular}{c|c}
    \midrule
    \textit{Target Dataset} & \textit{Single-Day kNN Accuracy}\\ \hline
    $2024-10-18$ & $98.96\%$\\
    $2024-10-31$ & $99.92\%$\\
    $2024-11-01$ & $97.57\%$\\
    $2024-11-02$ & $95.04\%$\\
    $2024-11-03$ & $95.30\%$\\
    $2024-11-04$ & $90.03\%$\\
    $2024-11-05$ & $95.36\%$\\
    $2024-11-06$ & $91.42\%$\\
    $2024-11-07$ & $90.36\%$\\
    \bottomrule
    $Avg+Std$ & \textbf{$94.88\pm3.63\%$}\\
    \end{tabular}
    \caption{\textbf{Single-Day UCL Accuracy} The kNN accuracies of the single-day UCL pipelines shown from Figure~\ref{fig:app-ex4}. Each instance trained and tested on data from the same day. $80\%$ of data is used for training and validation, and $20\%$ for testing.}
    \label{tab:UCLINF_table_SD}
\end{table}

\clearpage
\bibliographystyle{elsarticle-num}
\bibliography{ref}

\newpage






\end{document}